\documentclass[ijoc,sglanonrev]{informs4}
\usepackage{eqndefns-left} 
\RequirePackage{tgtermes}
\RequirePackage{newtxtext}
\RequirePackage{newtxmath}
\RequirePackage{bm}
\RequirePackage{endnotes}

\OneAndAHalfSpacedXII 

\usepackage{tikz}
\usepackage{multirow}
\usepackage[T1]{fontenc}
\usepackage{tgbonum}
\usepackage{graphicx}
\usepackage{subcaption}
\usepackage{amsmath,amssymb}
\usepackage{booktabs}
\usepackage{threeparttable}
\usepackage{color}  
\usepackage{fix-cm}
\usepackage{amsmath,amssymb}          
\usepackage{algorithm}                
\usepackage{algpseudocode}            
\usepackage{hyperref}

\newcommand{\sets}[1]{\ensuremath{\mathcal{#1}}}

\newcommand{\R}{\ensuremath{\mathbb{R}}} 

\usepackage{natbib}
\bibpunct[, ]{(}{)}{,}{a}{}{,}%
\def\bibfont{\small}%
\def\bibsep{\smallskipamount}%

\EquationsNumberedThrough    
\TheoremsNumberedThrough     
\ECRepeatTheorems

\MANUSCRIPTNO{IJOC-0001-2024.00}

\begin{document}

\RUNTITLE{Generalized  Optimal Classification Trees }

\TITLE{Generalized  Optimal Classification  Trees: A Mixed-Integer  Programming Approach}

\ARTICLEAUTHORS{%
\AUTHOR{Jiancheng Tu$^\text{a}$, Wenqi Fan$^\text{a, b}$, Zhibin Wu$^\text{c}$}
\AFF{{$^\text{a}$ Department of Computing, The Hong Kong Polytechnic University},  \EMAIL{jiancheng.tu@connect.polyu.hk}; 
$^\text{b}$ Department of Management and Marketing, The Hong Kong Polytechnic University, \EMAIL{wenqi.fan@polyu.edu.hk}; 
$^\text{c}$ Business School, Sichuan University, Chengdu 610065, P R China, \EMAIL{zhibinwu@scu.edu.cn}}}

\ABSTRACT{
Global optimization of decision trees is a long-standing challenge in combinatorial optimization, yet such models play an important role in interpretable machine learning. 
Although the problem has been investigated for several decades, only recent advances in discrete optimization have enabled practical algorithms for solving optimal classification tree problems on real-world datasets. 
Mixed-integer programming (MIP) offers a high degree of modeling flexibility, and we therefore propose a MIP-based framework for learning optimal classification trees under nonlinear performance metrics, such as the F1-score, that explicitly addresses class imbalance.
To improve scalability, we develop problem-specific acceleration techniques, including a tailored branch-and-cut algorithm, an instance-reduction scheme, and warm-start strategies.
We evaluate the proposed approach on 50 benchmark datasets. 
The computational results show that the framework can efficiently optimize nonlinear metrics  while achieving strong predictive performance and reduced solution times compared with existing  methods.
}


\maketitle

\section{Introduction}
Decision trees occupy a central role in interpretable machine learning, valued for their inherent transparency and ease of explanation. 
The classical Classification and Regression Trees (CART) algorithm~\citep{Breiman1984} exemplifies the dominant paradigm: tree structures are constructed via recursive, greedy splitting rules. 
Although such heuristics are computationally attractive, they provide no guarantee of global optimality, and the problem of finding the smallest or most accurate decision tree consistent with data is known to be NP-complete~\citep{laurent1976}. 
As a consequence, traditional practice has long relied on heuristic induction methods.

Recently, the global optimization of decision trees has attracted increasing attention, which seeks to construct tree models that are provably optimal with respect to prescribed objectives and constraints, rather than heuristic surrogates. 
A variety of exact approaches have been advanced, including mixed-integer programming (MIP)~\citep{Bertsimas2017,verwer2019,aghaei2025,firat2020,gunluk2021,Subramanian2023}, branch-and-bound algorithms~\citep{mazumder2022}, dynamic programming (DP)~\citep{linden2023,demirovic2023,Demirovic2022,Lin2020}, and  constraint programming (CP)~\citep{Shati2023,Nina2018}. 
These efforts have convincingly demonstrated that globally optimized trees can achieve superior predictive performance and model complexity compared to greedy induction.

Among these paradigms, MIP-based approaches stand out for their modeling power: they can naturally encode a wide range of objective functions and constraints. 
This flexibility has motivated continued research into MIP formulations for optimal classification trees and their scalable methods.

\subsection{MIP-based Approaches for Optimal Classification Trees}
\label{sec:related-work}

The foundational MIP model for optimal classification trees proposed by \citet{Bertsimas2017} showed that small-depth globally optimal trees could surpass classic heuristics such as CART, sparking an active line of research on improved formulations. 
Subsequent work has addressed issues of robustness~\citep{bertsimas2019},  stability~\citep{bertsimas2022stable}, and computational efficiency~\citep{verwer2019, gunluk2021, aghaei2025, firat2020, patel2024improved, Subramanian2023, alston2022}.

For instance, the \texttt{BinOCT} framework~\citep{verwer2019} leverages binary path-encoding variables to reduce model size, while \citet{aghaei2025} introduced a max-flow-based formulation (\texttt{FlowOCT}) and a Benders decomposition approach to improve tractability for larger problems. 
Other contributions have examined  cutting planes~\citep{michini2025polyhedral} and column generation methods~\citep{firat2020, patel2024improved, Subramanian2023}.

Despite these methodological advances, two core challenges persist: (i) most existing MIP frameworks can directly optimize only \emph{linear} objectives (e.g., misclassification error, balanced accuracy), and do not accommodate inherently nonlinear metrics such as the F1-score, Fowlkes–Mallows index, or Matthews correlation coefficient; and (ii) scalability remains a significant barrier, often requiring hours to solve medium-scale instances.

\subsection{Optimal Classification Trees for Imbalanced Data}
Class imbalance is prevalent in many high-stakes domains such as fraud detection~\citep{van2017gotcha} and medical diagnosis~\citep{chicco2020}. 
Most optimal classification tree formulations address imbalance by assigning class-specific misclassification weights or by maximizing balanced accuracy~\citep{Lin2020, linden2023, Shati2023, aghaei2025, tu2025}. 
While such linear objectives facilitate tractable MIP formulations, they may fail to fully reflect nonlinear metrics such as F1-score and the Matthews correlation coefficient.

A number of recent works have sought to incorporate these nonlinear metrics into optimal tree learning. 
For example, \citet{Subramanian2023} encode F1-score  as constraints, while \citet{tu2025} propose a mixed-integer quadratic programming (MIQP) model to directly maximize F1-score. However, the inherent non-convexity of these formulations imposes a severe computational burden.

Dynamic programming (DP) approaches~\citep{Lin2020, linden2023, demirovic2021} can optimize certain nonlinear metrics under specific conditions, but are restricted to feature sets and typically require the objective to be additively or separably structured. 
As a result, developing scalable methods for learning globally optimal trees under nonlinear, imbalance-aware metrics remains an open and challenging problem.

\subsection{Contributions}
To address the limitations identified above, we develop a unified MIP framework capable of directly optimizing a broad class of nonlinear, imbalance-aware metrics—including the F1-score, Matthews correlation coefficient (MCC), and Fowlkes–Mallows index—through tractable linearizations. 
We further design a problem-specific branch-and-cut algorithm that leverages problem structure and feature-aware instance compression to achieve superior scalability.
The principal contributions of this work are as follows:
\begin{itemize}
    \item We propose a general MIP-based formulation for learning optimal classification trees that can directly optimize a wide range of nonlinear metrics derived from the confusion matrix, offering a principled solution to imbalanced classification.
    \item We introduce a customized branch-and-cut algorithm that leverages feature-aware and conflict cuts to improve  scalability.
    \item Through extensive computational experiments on 50 benchmark datasets (ranging from 100 to 245,057 samples), we demonstrate that our framework consistently outperforms state-of-the-art methods such as BendersOCT~\citep{aghaei2025} in terms of scalability  and the dynamic programming approach of~\citet{demirovic2021} in terms of nonlinear objectives, delivering improved performance on imbalanced classification tasks and achieving substantial reductions in computation time.
\end{itemize}

Section~\ref{WeightedFlowOCT} establishes the core formulation. 
Section~\ref{sec:imbalanced} subsequently broadens the framework to handle nonlinear, imbalance-sensitive objective functions.
Section \ref{sec:scalable} introduces several acceleration techniques to reduce runtime.
Section~\ref{sec:experiments} reports computational results, and 
Section~\ref{sec:conclusion} concludes.

\section{Weighted Flow-based Optimal Classification Trees}
\label{WeightedFlowOCT}
We develop a weighted flow formulation for optimal classification trees that aggregates duplicate feature–label pairs into unique instances with integer multiplicities. 
Building on \texttt{FlowOCT} \citep{aghaei2025}, the resulting \texttt{WFlowOCT} preserves optimal solutions while reducing problem size. 
We solve it via a  Benders decomposition analogous to that employed in \texttt{BendersOCT} \citep{aghaei2025}.

\subsection{Problem Formulation}
\begin{figure}[]
\begin{center}
\centerline{
\includegraphics[width=0.35\textwidth]{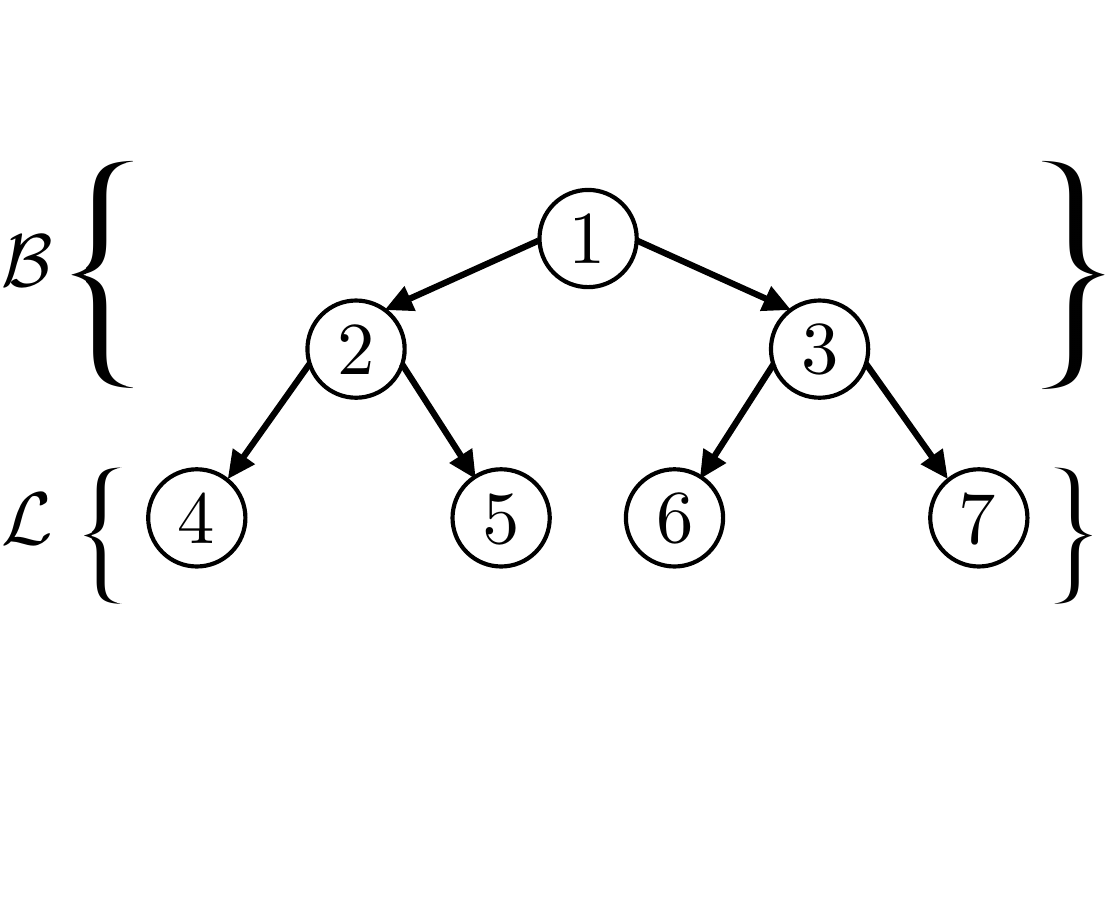}
\includegraphics[width=0.35\textwidth]{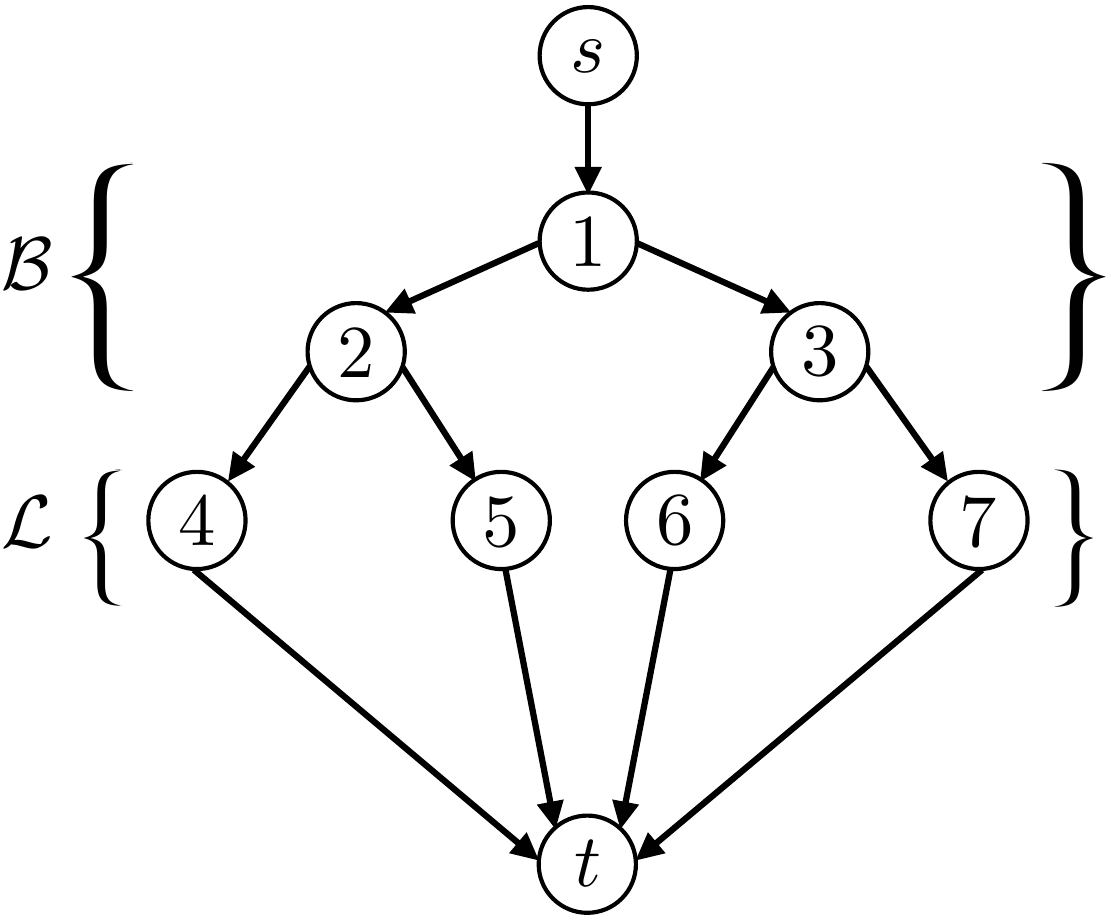}
}
\caption{A depth--2 decision tree (left) and its induced flow network (right).
The network is constructed by augmenting the tree nodes with a source and a sink; arcs correspond
to tree edges and terminal connections from leaves to the sink.
 Here, $\sets B=\{1,2,3\}$ and $\sets L=\{4,5,6,7\}$, while $\sets V=\{s,1,2,\ldots,7,t\}$ and $\sets A = \{(s,1),(1,2),\ldots,(7,t)\}$ \citep{aghaei2025}.}
\label{fig:sample_tree}
\end{center}
\end{figure}

Consider a classification problem defined over a feature index set $\mathcal{F}$ and a finite label set $\mathcal{K}$. 
The training data consist of binary-valued feature vectors paired with categorical labels, denoted by
$\mathcal{I}=\{(\mathbf{x}_i,y_i)\}_{i=1}^{|\mathcal{I}|}$, where $\mathbf{x}_i\in\{0,1\}^{|\mathcal{F}|}$ and
$y_i\in\{0,1,\ldots,|\mathcal{K}|-1\}$.
We represent the classifier as a complete binary decision tree of maximum depth $D$ with $T=2^{D+1}-1$ nodes, partitioned into branch nodes $\mathcal{B}=\{1,\dots,\lfloor T/2\rfloor\}$ and leaf nodes $\mathcal{L}=\{\lfloor T/2\rfloor+1,\dots,T\}$. 
Let $\mathcal{V}_{\mathrm{tree}}=\mathcal{B}\cup\mathcal{L}$ denote tree nodes. In the associated flow network (Figure~\ref{fig:sample_tree}), we use a source $s$ and a sink $t$, and define $\mathcal{V}=\{s\}\cup\mathcal{V}_{\mathrm{tree}}\cup\{t\}$.

\begin{definition}[Unique Dataset]
\label{def:unique-dataset}
Let $\mathcal{U}=\{(\mathbf{x}_i,y_i)\}_{i=1}^{|\mathcal{U}|}$ be the set of unique instances obtained from $\mathcal{I}$ by merging duplicates with identical $(\mathbf{x},y)$.  
Let $\mathcal{A}$ denote the set of post-discretization attributes, and for each $a\in\mathcal{A}$ let $m_a$ be the number of admissible bins or levels (for a binary attribute, $m_a=2$; for an attribute discretized into three bins, $m_a=3$).  
Then the number of unique instances satisfies
\begin{equation}
\label{eq:tight-undata}
|\mathcal{U}| \;\le\; \min\!\left\{\,|\mathcal{K}| \cdot \prod_{a\in\mathcal{A}} m_a,\; |\mathcal{I}|\,\right\}.
\end{equation}
For each unique instance $i\in\mathcal{U}$, let $w_i\in\mathbb{Z}_{\ge 1}$ denote its frequency in $\mathcal{I}$; then
\begin{equation}
\label{ui}
\sum_{i\in\mathcal{U}} w_i \;=\; |\mathcal{I}|.
\end{equation}
\end{definition}
The unique number of instances $|\mathcal{U}|$ is much lower than the upper bound determined by Equation (\ref{eq:tight-undata}). 
The results are presented in Table \ref{uniquetsummary}. For example, the dataset \textit{sepsis} contains 110,205 instances, but its unique dataset size is $|\mathcal{U}| = 140$, which is lower than its upper bound  32,768.

We adopt the standard \texttt{FlowOCT} graph (Figure~\ref{fig:sample_tree}) from \cite{aghaei2025}. 
For $n\in\mathcal{B}$, define the parent $a(n)=\lfloor n/2\rfloor$ (with $a(1)=s$), the left child $\ell(n)=2n$, and the right child $r(n)=2n+1$. For any node $n\in\mathcal{V}_{\mathrm{tree}}$, let $\mathcal{P}(n)$ denote its set of strict ancestors in $\mathcal{B}$.
Based on the unique dataset $\mathcal{U}$, we design the  weighted flow formulation (\texttt{WFlowOCT}).
We introduce the following decision variables: $b_{nf}\in\{0,1\}$ indicates feature $f$ is selected to split at branch $n$;
$p_n\in\{0,1\}$ indicates node $n\in\mathcal{V}_{\mathrm{tree}}$ is a terminal (leaf) node;
$c_{nk}\in\{0,1\}$ assigns label $k$ to node $n\in\mathcal{V}_{\mathrm{tree}}$;
$z^i_{a(n),n}$ indicates  that  data point $i$ is correctly classified and traverses the  arc (${a(n),n}$) (=1)   or not   $(= 0) $.
For a regularization parameter $\lambda\in[0,1)$, the \texttt{WFlowOCT} model is:
\begin{subequations}
\label{eq:flow_reg}
\begin{align}
\max\;\; & (1-\lambda)\sum_{i\in\mathcal{U}}\sum_{n\in\mathcal{V}_{\mathrm{tree}}} w_i\, z^i_{n,t}
\;-\; \lambda \sum_{n\in\mathcal{B}}\sum_{f\in\mathcal{F}} b_{nf} \label{eq:flow_reg_obj}\\
\text{s.t.}\;\;
& \sum_{f\in\mathcal{F}} b_{nf} \;+\; p_n \;+\; \sum_{m\in\mathcal{P}(n)} p_m \;=\;1 
&& \forall n\in\mathcal{B} \label{eq:flow_reg_branch_or_predict}\\
& p_n \;+\; \sum_{m\in\mathcal{P}(n)} p_m \;=\;1 
&& \forall n\in\mathcal{L} \label{eq:flow_reg_terminal_leaf}\\
& z^i_{a(n),n} \;=\; z^i_{n,\ell(n)} \;+\; z^i_{n,r(n)} \;+\; z^i_{n,t}
&& \forall n\in\mathcal{B},\;\forall i\in\mathcal{U} \label{eq:flow_reg_conservation_internal}\\
& z^i_{a(n),n} \;=\; z^i_{n,t}
&& \forall n\in\mathcal{L},\;\forall i\in\mathcal{U} \label{eq:flow_reg_conservation_terminal}\\
& z^i_{s,1} \;\le\; 1
&& \forall i\in\mathcal{U} \label{eq:flow_reg_source}\\
& z^i_{n,\ell(n)} \;\le\; \sum_{f\in\mathcal{F}: x_f^i=0} b_{nf}
&& \forall n\in\mathcal{B},\;\forall i\in\mathcal{U} \label{eq:flow_reg_branch_left}\\
& z^i_{n,r(n)} \;\le\; \sum_{f\in\mathcal{F}: x_f^i=1} b_{nf}
&& \forall n\in\mathcal{B},\;\forall i\in\mathcal{U} \label{eq:flow_reg_branch_right}\\
& z^i_{n,t} \;\le\; c_{n,y_i}
&& \forall n\in\mathcal{V}_{\mathrm{tree}},\;\forall i\in\mathcal{U} \label{eq:flow_reg_sink}\\
& \sum_{k\in\mathcal{K}} c_{nk} \;=\; p_n
&& \forall n\in\mathcal{V}_{\mathrm{tree}} \label{eq:flow_reg_leaf_prediction}\\
& b_{nf},\, p_n,\, c_{nk},\, z^i_{a(n),n},\, z^i_{n,t} \in \{0,1\}
&& 
\end{align}
\end{subequations}

The main differences between \texttt{WFlowOCT} and \texttt{FlowOCT} lie in their objective functions and in the range of the decision variables $z^i_{a(n),n}$.
The constraints of \texttt{WFlowOCT} are structurally similar to those of \texttt{FlowOCT}; detailed formulations and explanations of the original \texttt{FlowOCT} constraints are provided in E-Companion~\ref{appendix_sec: FlowOCT}.
A key computational benefit of \texttt{WFlowOCT} arises when the number of unique
instances satisfies $|\mathcal{U}|<|\mathcal{I}|$, in which case the size of the mixed-integer
formulation is dramatically reduced.
For instance, for the \texttt{sepsis} dataset reported in Table~\ref{uniquetsummary},
the ratio $|\mathcal{I}|/|\mathcal{U}| = 110{,}204/140 \approx 787$ indicates that the
flow-assignment components of \texttt{WFlowOCT} scale by only about $1/787$ relative to
their counterparts in \texttt{FlowOCT}.

\begin{proposition}
\label{Proposition:WFlowOCTeFlowOCT}
\texttt{WFlowOCT} is equivalent to \texttt{FlowOCT} in the sense that there exists a bijection between optimal solutions:
given any optimal \texttt{FlowOCT} solution on $\mathcal{I}$, there is an optimal \texttt{WFlowOCT} solution on $(\mathcal{U},w)$ with identical tree structure and objective value, and vice versa.
\end{proposition}

Following \citet{aghaei2025}, we adopt a Benders decomposition tailored to \texttt{WFlowOCT}. 
Let $g^{i}(\bm b,\bm c,\bm p)$ denote the maximum feasible $s$--$t$ flow for instance $i\in\mathcal{U}$ under capacities induced by $(\bm b,\bm c,\bm p)$. 
The master problem (MP) optimizes over the structural decision variables $(\bm b,\bm c,\bm p)$, while the subproblem (SP) computes each $g^i$ as a max-flow on the fixed network associated with instance~$i$.

Solving the SP implies a family of standard max-flow Benders cuts. 
For each instance $i\in\mathcal{U}$ and each $s$-separating cut $\mathcal{S}\subseteq\mathcal{V}\setminus\{t\}$ with $s\in\mathcal{S}$, let $\mathcal{C}(\mathcal{S})$ denote the set of arcs crossing the cut. 
Introducing variables $g^i\in \{0,1\}$ to represent the maximum attainable flow for instance $i$, the resulting Benders master problem is:
\begin{subequations}
\label{eq:bendersOCT_master_2}
\begin{align}
\max\;\; & (1-\lambda)\sum_{i\in\mathcal{U}} w_i\, g^i 
          \;-\; \lambda \sum_{n\in\mathcal{B}}\sum_{f\in\mathcal{F}} b_{nf} 
          \label{eq:bendersOCT_master_2_obj}\\[2pt]
\text{s.t.}\;\;
& g^i
   \;\le\; 
   \sum_{(n_1,n_2)\in \mathcal{C}(\mathcal{S})} 
        c^i_{n_1,n_2}(\bm b,\bm c)
   && \forall i\in\mathcal{U},\;\forall \mathcal{S}\subseteq\mathcal{V}\setminus\{t\}: s\in\mathcal{S} 
   \label{eq:bendersOCT_master_2_benders_cut1}\\[2pt]
& \eqref{eq:flow_reg_branch_or_predict},\;
  \eqref{eq:flow_reg_terminal_leaf},\;
  \eqref{eq:flow_reg_leaf_prediction},\qquad
  b_{nf},\, p_n,\, c_{nk}\in\{0,1\} \nonumber\\[2pt]
& g^i\in \{0,1\} 
   && \forall i\in\mathcal{U}.
   \label{eq:bendersOCT_master_2_g_upperbound}
\end{align}
\end{subequations}

We refer to \eqref{eq:bendersOCT_master_2} as \texttt{BendersOCT-weighted}.  
The Benders cuts in \eqref{eq:bendersOCT_master_2_benders_cut1} are added dynamically through the Gurobi callback.  
We further observe that the separation algorithm of \citet{aghaei2025} can be directly applied to identify violated cuts efficiently.

\begin{proposition}
\label{Proposition:eq:bendersOCT_master_2}
Separation of the Benders cuts in \eqref{eq:bendersOCT_master_2_benders_cut1} follows the min-cut routine described in Algorithm~2 of \citet{aghaei2025} (refer to Algorithm  \ref{alg:Separation} in E-Companion).
\end{proposition}

\section{Imbalanced Datasets}
\label{sec:imbalanced}
In many real-world applications, the positive class is rare, and standard accuracy may be misleading.  
This section demonstrates that our framework accommodates both linear and nonlinear imbalance-aware objectives within a unified MIP architecture based on the \texttt{BendersOCT} structure.  
All objectives considered are functions of the confusion-matrix entries \(\mathrm{TP},\mathrm{TN},\mathrm{FP},\mathrm{FN}\).  
The proposed framework is distinguished by its modeling flexibility: any metric expressible as a  function of these integer counts with a nonnegative denominator can be embedded within the MIP.  
Given a performance metric \(h(\mathrm{TP},\mathrm{TN},\mathrm{FP},\mathrm{FN})\), if \(h\) is nonlinear, we reformulate it using fractional programming, binary expansion, and McCormick envelopes.  
This yields a \emph{unified} MIP model applicable to a wide family of metrics—including \(F_\beta\), MCC, balanced accuracy, G-Mean, Fowlkes--Mallows, and Jaccard/IoU—without altering the underlying structure.

We detail two representative nonlinear metrics: the \(F_\beta\) score (a linear-fractional measure) and the Matthews correlation coefficient (MCC, a correlation-type measure).  
Both are widely used in imbalanced classification.  
Formulations for other metrics (e.g.,  G-Mean, and Fowlkes--Mallows) follow the same pattern and appear in E-Companion~\ref{app:other-metrics}.

We consider binary classification (\(|\mathcal{K}|=2\), labels \(\{0,1\}\)).  
Let 
\[
n^+ = \sum_{i\in\mathcal{U}:y_i=1} w_i, \qquad
n^- = \sum_{i\in\mathcal{U}:y_i=0} w_i.
\]
Then the confusion-matrix entries are
\begin{align}
\mathrm{TP} &= \sum_{i\in\mathcal{U}:y_i=1} w_i g^i, &
\mathrm{FN} &= n^+ - \sum_{i\in\mathcal{U}:y_i=1} w_i g^i, \label{eq:tp-fn}\\
\mathrm{TN} &= \sum_{i\in\mathcal{U}:y_i=0} w_i g^i, &
\mathrm{FP} &= n^- - \sum_{i\in\mathcal{U}:y_i=0} w_i g^i. \label{eq:tn-fp}
\end{align}

\subsection{\texorpdfstring{$F_\beta$}{F-beta} Score}

The \(F_\beta\) score, including the widely adopted F1-score, is frequently used to evaluate binary classifiers  when the dataset is imbalanced:
\begin{equation}
\label{eq:f_beta_def}
F_\beta 
\;=\;
\frac{(1+\beta^2)\operatorname{Precision}\cdot \operatorname{Recall}}
{\beta^2\operatorname{Precision} + \operatorname{Recall}}
\;=\;
\frac{(1+\beta^2)\cdot \mathrm{TP}}
{\beta^2 n^+ + n^- + \mathrm{TP} - \mathrm{TN}},
\end{equation}
where $\mathrm{TP}$ and$\mathrm{TN}$  denote true positives and true negatives, respectively.

Recent optimal classification tree models have sought to directly maximize F1-score \citep{tu2025, tu2024, linden2023, demirovic2021, Lin2020}, but their scalability remains limited.  
In our prior work \citep{tu2025, tu2024}, we formulated the problem as a nonconvex MIQP, which becomes intractable for larger datasets or deeper trees.  
Here, we reformulate it as a pure MIP by linearizing the single bilinear term coupling \(F_\beta\) with confusion-matrix entries, thereby achieving substantial scalability gains (see Section~\ref{ref:experiment sub_Linearization}).

Let \(F_\beta\in[0,1]\) be a decision variable representing the score.  
A direct MIQP formulation is:
\begin{subequations}
\label{MIQP_Fbeta}
\begin{align}
\max\;\; & F_\beta - \lambda \sum_{n\in\mathcal{B}}\sum_{f\in\mathcal{F}} b_{nf} \label{eq:miqpf_obj}\\
\text{s.t.}\;\;
& F_\beta \cdot ({\beta^2 n^+ + n^- +\mathrm{TP} - \mathrm{TN}}) \;\leq\; {(1+\beta^2) \cdot \mathrm{TP}}, \label{eq:miqpf_frac}\\
& g^i \;\le\; \sum_{(n_1,n_2)\in \mathcal{C}(\mathcal{S})} c^i_{n_1,n_2}(\bm b,\bm c)
&& \forall i\in\mathcal{U},\;\forall \mathcal{S}\subseteq\mathcal{V}\setminus\{t\}: s\in\mathcal{S}, \label{eq:bendersOCT_master_2_benders_cut}\\
& \eqref{eq:flow_reg_branch_or_predict}, \eqref{eq:flow_reg_terminal_leaf}, \eqref{eq:flow_reg_leaf_prediction},\quad
b_{nf},p_n,c_{nk}\in\{0,1\}, \\
& g^i \in \{0, 1\} && \forall i\in\mathcal{U}. \label{eq:bendersOCT_master_2_g_upperbound}
\end{align}
\end{subequations}

The only nonlinearity appears in constraint~\eqref{eq:miqpf_frac}.
Let \(l=\left\lceil \log_2|\mathcal{I}|\right\rceil\).
Introduce binary variables \(\delta_k\) for \(k=0,\dots,l\) and impose
$
n^{-} + \mathrm{TP} - \mathrm{TN} \;=\; \sum_{k=0}^{l} 2^{k}\delta_k$.
Multiplying both sides by \(F_\beta\) yields
$F_\beta(n^{-} + \mathrm{TP} - \mathrm{TN}) \;=\; \sum_{k=0}^{l} 2^{k}F_\beta\delta_k$.
Introduce auxiliary variables \(\gamma_k\) for \(k=0,\dots,l\) such that 
\( \gamma_k = F_\beta\delta_k\).
Using McCormick envelopes (which are exact on \([0,1]\times\{0,1\}\)), we impose
$\gamma_k \le \delta_k,\quad
\gamma_k \le F_\beta,\quad
\gamma_k \ge F_\beta + \delta_k - 1$.
The model \eqref{MIQP_Fbeta} can thus be transformed into the following MIP model:
\begin{subequations}
\label{eq:OCTF-MIP}
\begin{align}
\max\;\; & F_\beta - \lambda \sum_{n\in\mathcal{B}}\sum_{f\in\mathcal{F}} b_{nf} \label{eq:mipf_obj}\\
\text{s.t.}\;\;
&   F_\beta(\beta^2 n^+ + n^-) + \sum_{k=0}^{l} 2^k \,\gamma_k \;\leq\; (1+\beta^2)\sum_{i\in\mathcal{U}: y_i=1} w_i g^i, \label{eq:F1_con}\\
& g^i \;\le\; \sum_{(n_1,n_2)\in \mathcal{C}(\mathcal{S})} c^i_{n_1,n_2}(\bm b,\bm c)
&& \forall i\in\mathcal{U},\;\forall \mathcal{S}\subseteq\mathcal{V}\setminus\{t\}: s\in\mathcal{S}, \label{eq:bendersOCT_master_2_benders_cut_F}\\
& \eqref{eq:flow_reg_branch_or_predict}, \eqref{eq:flow_reg_terminal_leaf}, \eqref{eq:flow_reg_leaf_prediction},\quad
b_{nf},p_n,c_{nk}\in\{0,1\}, \\
&  n^{-} + \mathrm{TP} - \mathrm{TN} =\sum_{k=0}^{l} 2^{k}\delta_k,  \label{eq:F1_con1}\\
&\gamma_k \le \delta_k,\quad
\gamma_k \le F_\beta,\quad
\gamma_k \ge F_\beta + \delta_k - 1 && \forall k, \\
& g^i \in \{0, 1\} && \forall i\in\mathcal{U}. \label{eq:gi}
\end{align}
\end{subequations}

\begin{remark}
The linearization introduces \(l+1 = \left\lceil \log_2|\mathcal{I}|\right\rceil+1\) new binary variables \(\delta_k\),
the same number of continuous variables \(\gamma_k\), and at most \(3(l+1)\) linear constraints from the McCormick envelopes.
Thus the additional model size is logarithmic in the number of (weighted) samples,
and in practice the contribution of these variables and constraints is negligible compared to
the structural variables \((b,p,c,g)\) and the Benders cuts.
This explains why the \(F_\beta\) formulation scales similarly to the accuracy-based model in our experiments.
\end{remark}

\subsection{Matthews Correlation Coefficient (MCC)}
\label{subsec:mcc}

The Matthews correlation coefficient (MCC) is a correlation-based measure of classification quality that remains reliable under class imbalance:
\begin{equation}
\begin{aligned}
\mathrm{MCC} &= \frac{\mathrm{TP} \cdot \mathrm{TN} \;-\; \mathrm{FP} \cdot \mathrm{FN}}{\sqrt{(\mathrm{TP+FP})\,(\mathrm{TP+FN})\,(\mathrm{TN+FP})\,(\mathrm{TN+FN})}} \\
&=\frac{A}{\sqrt{n^+ n^- \, U \, V}},
\end{aligned}
\label{eq:MCC-def}
\end{equation}
where we define
\[
A = n^+ \mathrm{TN} + n^- \mathrm{TP} - n^+ n^-,\qquad
U = \mathrm{TP}-\mathrm{TN}+ n^-,\qquad
V = \mathrm{TN}-\mathrm{TP}+ n^+.
\]

\begin{proposition}
\label{Proposition:Apositive}
For any binary dataset, maximizing the MCC is equivalent to maximizing the square of the MCC in the FlowOCT structure.
\end{proposition}

Motivated by Proposition~\ref{Proposition:Apositive}, we introduce a decision variable 
\(\mathrm{MCC2}\in[0,1]\) representing \(\mathrm{MCC}^2\) and obtain the following non-convex MIQCP formulation for maximizing the MCC in the BendersOCT structure:

\begin{equation}
\label{eq:OCT-MCC2-MIQCP}
\begin{array}{llll}
\max & \mathrm{MCC2} \\[2pt]
\text{s.t.} 
& \displaystyle n^+ n^- \cdot \mathrm{MCC2}\cdot S \;\le\; A^2, \\[2pt]
& A \;=\; n^+ \mathrm{TN} + n^- \mathrm{TP} - n^+ n^-,\\
& U \;=\; \mathrm{TP}-\mathrm{TN}+ n^-,\quad V \;=\; \mathrm{TN}-\mathrm{TP}+ n^+,\\
& S = U V, \\
& A \geq 0,\\
& 0\le \mathrm{MCC2}\le 1,\\
& \text{Constraints\ \eqref{eq:tp-fn}--\eqref{eq:tn-fp}, \eqref{eq:bendersOCT_master_2_benders_cut}--\eqref{eq:bendersOCT_master_2_g_upperbound}.}
\end{array}
\end{equation}

In model \eqref{eq:OCT-MCC2-MIQCP}, the nonlinear terms \(\mathrm{MCC2}\cdot (UV)\) and \(A^2\) make the model non-convex and hard to solve. 
We therefore linearize \(\mathrm{MCC2}\cdot (UV)\) and \(A^2\) via binary expansions and McCormick envelopes.

\paragraph{(i) Binary expansion of \(U\cdot V\).}
Since \(0\le U,V\le n^++n^- = |\mathcal{I}|\), we need at most
$L \;=\; \left\lceil \log_2|\mathcal{I}|\right\rceil
$
binary digits to represent each of \(U\) and \(V\).
Introduce binary variables \(u_r,v_s\) for \(r,s=0,\dots,L\) and write
\begin{equation}
U \;=\; \sum_{r=0}^{L} 2^{r}\,u_r,\qquad
V \;=\; \sum_{s=0}^{L} 2^{s}\,v_s.
\label{eq:U-V-bits}
\end{equation}
Then introduce binary variables \(\chi_{rs}\) enforcing \(\chi_{rs} = u_r\land v_s\), and obtain
\begin{equation}
U V \;=\; \sum_{r=0}^{L}\sum_{s=0}^{L} 2^{r+s}\,\chi_{rs},
\end{equation}
with constraints:
\begin{equation}
\chi_{rs}\le u_r,\quad \chi_{rs}\le v_s,\quad \chi_{rs}\ge u_r+v_s-1
\qquad \forall r,s.
\label{eq:UV-prod}
\end{equation}

\paragraph{(ii) Linearization of \(\mathrm{MCC2}\cdot (U V)\).}
Define \(\theta_{rs}=\mathrm{MCC2}\cdot \chi_{rs}\) and impose McCormick envelopes:
\begin{equation}
\theta_{rs}\le \chi_{rs},\quad \theta_{rs}\le \mathrm{MCC2},\quad \theta_{rs}\ge \mathrm{MCC2}-(1-\chi_{rs})\qquad \forall r,s,
\label{eq:MCC2-chi}
\end{equation}
which implies
\begin{equation}
\mathrm{MCC2}\,(U V) \;=\; \sum_{r=0}^{L}\sum_{s=0}^{L} 2^{r+s}\,\theta_{rs}.
\label{eq:MCC2UV}
\end{equation}

\noindent(iii) \emph{Linearize the quadratic term \(A^2\).} 
Since \(|A|\le n^+ n^-\), let 
$
T = \left\lceil \log_2(n^+ n^-)\right\rceil$.
Introduce binary variables \(\beta_t\in\{0,1\}\) for \(t=0,\dots,T\) such that
\begin{equation}
A \;=\; \sum_{t=0}^{T} 2^{t}\,\beta_t. 
\label{eq:A-bits}
\end{equation}
With ``AND'' binaries \(\pi_{tt'}\in\{0,1\}\) enforcing \(\pi_{tt'}=\beta_t\land \beta_{t'}\), we obtain 
\begin{equation}
A^2 \;=\; \sum_{t=0}^{T}\sum_{t'=0}^{T} 2^{t+t'}\,\pi_{tt'},
\label{eq:A-sq}
\end{equation}
with constraints:
\begin{equation}
\pi_{tt'}\le \beta_t,\quad 
\pi_{tt'}\le \beta_{t'},\quad 
\pi_{tt'}\ge \beta_t+\beta_{t'}-1
\qquad \forall t,t'.
\label{eq:A-sqc}
\end{equation}

Replacing the nonlinear terms \(\mathrm{MCC2}\cdot (UV)\) and \(A^2\) in model (\ref{eq:OCT-MCC2-MIQCP}) with the linear expressions \eqref{eq:MCC2UV} and \eqref{eq:A-sq}, we obtain the following MIP formulation:

\begin{equation}
\label{eq:OCT-MCC2-MIP}
\begin{array}{rlll}
\max & \mathrm{MCC2} \\[2pt]
\text{s.t.} 
& \displaystyle 
  n^+ n^- \sum_{r=0}^{L}\sum_{s=0}^{L} 2^{r+s}\,\theta_{rs} 
  \;\le\; 
  \sum_{t=0}^{T}\sum_{t'=0}^{T} 2^{t+t'}\,\pi_{tt'},  \\[6pt]
& A \;=\; n^+ \mathrm{TN} + n^- \mathrm{TP} - n^+ n^-,\\
& U \;=\; \mathrm{TP}-\mathrm{TN}+ n^-,\quad V \;=\; \mathrm{TN}-\mathrm{TP}+ n^+,\\
& U \;=\; \sum_{r=0}^{L} 2^{r}\,u_r,\qquad V \;=\; \sum_{s=0}^{L} 2^{s}\,v_s, \\[2pt]
& 0\le \mathrm{MCC2}\le 1,\quad A \geq 0,\\[2pt]
&\text{Constraints\ \eqref{eq:tp-fn}--\eqref{eq:tn-fp}, 
 \eqref{eq:bendersOCT_master_2_benders_cut}--\eqref{eq:bendersOCT_master_2_g_upperbound}, 
  \eqref{eq:UV-prod}--\eqref{eq:MCC2-chi}, \eqref{eq:A-sqc}.}
\end{array}
\end{equation}

\begin{remark}
The MCC linearization adds ${2(L+1) + (L+1)^2 + (T+1) + (T+1)^2}$ binary variables and ${3(L+1)^2 + 3(L+1)^2 + 3(T+1)^2}$ linear constraints.
Hence both the number of auxiliary \(0\)–\(1\) variables and the number of additional constraints grow 
polylogarithmically in the sample size \(|\mathcal{I}|\).
The constant factors are larger than for the \(F_\beta\) formulation, 
but for moderate-depth trees and UCI-scale datasets this overhead remains manageable and the resulting MIP is tractable.
For very large training sets or substantially deeper trees, the MCC formulation can become heavy, 
and using lighter objectives such as \(F_\beta\)  may be more practical.
\end{remark}

\subsection{Combinations}

Following the same template as for the \(F_\beta\) score and MCC, we can derive MIQP and MIP formulations for any metric computed from the confusion table.
Appendix~\ref{app:other-metrics} provides formulations for other commonly used performance metrics.

In binary classification, decision makers rarely choose a model based on a single performance indicator (e.g., accuracy or F1-score), because different metrics emphasize different trade-offs and may rank models inconsistently, especially under class imbalance. 
Instead, a practical strategy is to select a \emph{stable} classifier that performs competitively across a set of metrics---such as F1-score, balanced accuracy (BA), and MCC. 
Our BendersOCT architecture makes it straightforward to maximize any convex combination of metrics constructed from the confusion matrix.
For example, to maximize both accuracy and \(F_\beta\), we consider the MIP model:
\begin{subequations}
\label{eq:MIP-acc_f}
\begin{align}
\max \quad 
& \alpha_1 F_\beta 
  + \frac{\alpha_2}{|\mathcal{I}|} \sum_{i\in\mathcal{U}} w_i\, g^i 
  - \lambda \sum_{n\in\mathcal{B}} \sum_{f\in\mathcal{F}} b_{nf}, \label{eq:MIP-acc_f-obj} \\[4pt]
\text{s.t.} \quad 
& \text{Constraints }(\ref{eq:F1_con})\text{--}(\ref{eq:gi}), \label{eq:MIP-acc_f-const}
\end{align}
\end{subequations}
where \(\alpha_1, \alpha_2 \geq 0\) control the relative importance of \(F_\beta\) and accuracy.

\section{Acceleration Techniques}
\label{sec:scalable}

We now describe two complementary mechanisms that improve  scalability of our formulations:
(i) cutting-plane techniques based on conflict subsets and feature-activated inequalities; and
(ii) a feasible-solution injection framework that combines depth-incremental warm starts with
LP-guided heuristic injections at MIP nodes.

\subsection{Cutting Planes}
\label{subsec:cutting-plane}

\subsubsection{Valid inequalities}

We first exploit the fact that duplicated feature vectors with different labels cannot all be
classified correctly by any decision tree.

\begin{definition}[Conflict subset]
\label{def:conflict_subset}
A subset $\mathcal{G}_s \subseteq \mathcal{U}$ is a \emph{conflict subset} if it contains instances
with identical features but non-identical labels.
For each class \(k\in\mathcal{K}\), define
\(\mathcal{G}_{sk}=\{i\in\mathcal{G}_s:\,y_i=k\}\), so that
\(\mathcal{G}_s=\bigcup_{k\in\mathcal{K}}\mathcal{G}_{sk}\) and
\(|\mathcal{G}_s|=\sum_{k\in\mathcal{K}}|\mathcal{G}_{sk}|\).
\end{definition}

If all points in \(\mathcal{G}_s\) share the same feature vector, any decision tree induces the same
path and prediction for all of them. Hence at most one class in \(\mathcal{G}_s\) can be classified
correctly.

\begin{theorem}
\label{thm:conflict_subset_improvement}
For any conflict subset \(\mathcal{G}_s\), the following inequality is valid for
\texttt{BendersOCT-weighted}:
\begin{equation}
\label{equation:lowers}
\sum_{i\in\mathcal{G}_s} g^i \;\le\; \max_{k\in\mathcal{K}} |\mathcal{G}_{sk}|.
\end{equation}
\end{theorem}

\begin{example}
\label{ex:conflict_toy}
Consider the following example with \(\mathcal{K}=\{0,1\}\),
four instances \(\mathcal{U}=\{1,2,3,4\}\), and four features
\(\mathcal{F}=\{1,2,3,4\}\), with unit weights \(w_i=1\) and
\[
\mathbf{x}_1 = (0,1,0,0)^\top,\ y_1=0;\quad
\mathbf{x}_2 = (0,1,0,0)^\top,\ y_2=1;
\]
\[
\mathbf{x}_3 = (1,0,1,0)^\top,\ y_3=0;\quad
\mathbf{x}_4 = (1,0,1,1)^\top,\ y_4=1.
\]
Under the full feature set \(\mathcal{F}\), the only conflict subset is
\(\mathcal{G}_1=\{1,2\}\).
We have
\(\mathcal{G}_{10}=\{1\}\), \(\mathcal{G}_{11}=\{2\}\),
so \(\max_{k}|\mathcal{G}_{1k}|=1\) and
\[
g^1 + g^2 \;\le\; 1,
\]
which encodes that instances \(1\) and \(2\) must share a prediction path and thus cannot both be correct.
\end{example}

The inequalities in Theorem~\ref{thm:conflict_subset_improvement} extend naturally to
\emph{feature-activated} conflicts that arise when a subset of features is disabled.

\begin{proposition}[Feature-activated conflict subset]
\label{prop:Fs}
Let \(\mathcal{F}'\subseteq\mathcal{F}\) and suppose that, when features in \(\mathcal{F}'\) are
removed (i.e., the tree cannot branch on them), a collection of instances merges into a conflict
subset \(\mathcal{G}_s'\subseteq\mathcal{U}\).
Partition \(\mathcal{G}_s'\) by class as
\(\mathcal{G}'_{sk}=\{i\in\mathcal{G}_s':\,y_i=k\}\), \(k\in\mathcal{K}\).
Then the following inequality is valid for \texttt{BendersOCT-weighted}:
\begin{equation}
\label{eq:Fs}
\sum_{i\in\mathcal{G}_s'} g^i
\;\le\;
\max_{k\in\mathcal{K}} |\mathcal{G}'_{sk}|
\;+\;
\Bigl(|\mathcal{G}_s'|-\max_{k\in\mathcal{K}}|\mathcal{G}'_{sk}|\Bigr)
\sum_{n\in\mathcal{B}}\sum_{f\in\mathcal{F}'} b_{nf}.
\end{equation}
\end{proposition}

Inequality~\eqref{eq:Fs} interpolates between two regimes:
if \(\sum_{n,f\in\mathcal{F}'} b_{nf}=0\), the right-hand side reduces to
\(\max_k|\mathcal{G}'_{sk}|\), recovering the pure conflict bound;
if \(\sum_{n,f\in\mathcal{F}'} b_{nf}\ge 1\), the right-hand side becomes \(|\mathcal{G}_s'|\),
and the inequality is automatically satisfied.
Thus, \eqref{eq:Fs} prevents the LP relaxation from simultaneously (i) assigning
\(g^i\approx 1\) for all \(i\in\mathcal{G}'_s\) and (ii) keeping features in \(\mathcal{F}'\)
fractionally unused.

\begin{example}
Reconsider Example~\ref{ex:conflict_toy} with \(\mathcal{F}'=\{4\}\).
In the reduced feature space \(\{1,2,3\}\),
\[
\mathbf{x}_3|_{\{1,2,3\}} = (1,0,1)^\top,\quad
\mathbf{x}_4|_{\{1,2,3\}} = (1,0,1)^\top,
\]
so instances \(3\) and \(4\) form a feature-activated conflict subset
\(\mathcal{G}'_1=\{3,4\}\) with
\(\mathcal{G}'_{10}=\{3\}\), \(\mathcal{G}'_{11}=\{4\}\),
\(|\mathcal{G}_1'|=2\), and \(\max_k|\mathcal{G}'_{1k}|=1\).
Inequality~\eqref{eq:Fs} becomes
\[
g^3 + g^4 \;\le\; 1 + \sum_{n\in\mathcal{B}} b_{n4}.
\]
If \(\sum_{n}b_{n4}=0\), at most one of \(\{3,4\}\) can be correct; if some split on feature~4 is
used, the bound is \(\ge 2\) and the inequality is nonbinding.
\end{example}

\subsubsection{Branch-and-cut implementation}
\label{subsec:b&c}

The number of pure conflict subsets \(\mathcal{G}_s\) is typically small, so the inequalities
\eqref{equation:lowers} are added as static constraints to the Benders master.
In contrast, feature-activated conflict subsets \(\mathcal{G}_s'\) depend on the subset
\(\mathcal{F}'\) of features that are effectively ``switched off'' in the LP relaxation and are
separated dynamically.

Let \((\mathbf{b}^{*}, \mathbf{g}^{*})\) be the LP solution at a given  node.
We observe empirically that most relaxations satisfy \(g^{*}_{i} \approx 1\) for most
\(i \in \mathcal{U}\), and many features are essentially unused, that is, for most of the features $\sum_{n \in \mathcal{B}} b^{*}_{nf} \approx 0$.
We therefore define
\[
\mathcal{F}' \;=\; \Bigl\{ f \in \mathcal{F} \,\Big|\, \sum_{n \in \mathcal{B}} b^{*}_{nf} = 0 \Bigr\},
\]
group instances that share the same pattern on \(\mathcal{F}\setminus\mathcal{F}'\) into
subsets \(\mathcal{G}'_s\), and check each for violation of~\eqref{eq:Fs}.
Whenever a violation is found, the corresponding cut is added via the solver's callback function.
These cuts depend only on \((\mathbf{b},\mathbf{g})\), and thus can be applied to Benders master
\eqref{eq:bendersOCT_master_2} and all its variants (linear and nonlinear objectives).
We refer to the resulting master with conflict and feature-activated cuts as
\texttt{BendersOCT-cut} in our experiments.

\subsection{Feasible-Solution Injection}
\label{subsec:feasible-injection}

From a branch-and-bound perspective, a strong initial incumbent immediately tightens the global upper bound, 
shrinks the portion of the search space that must be explored, and improves node selection and pruning efficiency \citep{Morrison2016}.
Prior work has shown that injecting the CART solution can quickly produce a reasonable integer-feasible tree and reduce solution time \citep{tu2025, Bertsimas2017}.  
However, CART is a heuristic and its solution quality can be limited especially for
deeper trees or nonlinear objectives.
We therefore design a unified feasible-solution injection framework that enhances the scalability of
\texttt{BendersOCT}. It relies on three empirical observations:
\begin{enumerate}
    \item If the MIP only considers the features used by CART, then \texttt{BendersOCT} often improves on the CART objective
    at the same depth.
    \item Optimal classification trees typically use only a few features (often \(\le 10\)), and shallow trees with small
    feature sets can be solved quickly.
    \item Any shallow tree is a feasible subtree of deeper trees with the same branching pattern; thus,
    solutions at smaller depths can be lifted to provide warm starts for deeper models.
\end{enumerate}

These observations motivate a three-layer strategy:
(i) a depth-incremental warm start from depth~\(2\);
(ii) a global Random Forest (RF) ranking that guides the evolution of feature sets;
and (iii) an LP-guided node heuristic that injects additional incumbents during the MIP search.

\subsubsection{Depth-incremental warm start and RF-guided feature sets}

We solve a sequence of OCT problems for depths \(d = 2,\dots,D\), using a single global RF ranking.
Let \(T_d\) denote the depth-\(d\) tree and \(\mathcal{F}_d\) its feature set.
The   procedure is detailed Algorithm \ref{alg:feasible-injection}.

First, fit a CART of depth \(D\) and let
$\mathcal{F}_{\mathrm{CART}} = \{f \in \mathcal{F} \mid f \text{ is used in CART}\}.
$
Next, fit an RF and compute importance scores \(I_f\) for \(f\in\mathcal{F}\), then obtain the global
ranking
$
L = \bigl(f^{(1)}, f^{(2)},\dots,f^{(p)}\bigr)
\quad\text{such that}\quad
I_{f^{(1)}} \ge I_{f^{(2)}} \ge \cdots \ge I_{f^{(p)}}$.

For \(d=2\), we set
$
\mathcal{F}_2 \leftarrow \mathcal{F}_{\mathrm{CART}}
\cup \{ \text{first } k \text{ indices in } L \setminus \mathcal{F}_{\mathrm{CART}}\}$.
We restrict the data to \(X_{\mathcal{F}_2}\), solve the OCT MIP on
\((X_{\mathcal{F}_2},y)\) to obtain a warm start \(S^{\mathrm{warm}}_2\), and then solve the full
depth-\(2\) MIP on \((X,y)\) with this warm start and node-level heuristics, yielding \(T_2\) and an
updated \(\mathcal{F}_2\).

For \(d>2\), we iterate
$\mathcal{F}_d \leftarrow \mathcal{F}_{d-1}
\cup \{ \text{next } k \text{ indices in } L \setminus \mathcal{F}_{d-1} \}$,
embed \(T_{d-1}\) into a depth-\(d\) topology over \(\mathcal{F}_d\) to form a reduced warm start
\(S^{\mathrm{sub}}_d\), solve the reduced problem on \((X_{\mathcal{F}_d},y)\) to obtain
\(\tilde{T}_d\), and lift \(\tilde{T}_d\) back to the full space to construct a full warm start
\(S^{\mathrm{full}}_d\).
The depth-\(d\) OCT MIP on \((X,y)\) is then re-optimized with \(S^{\mathrm{full}}_d\) and node-level
heuristics to obtain \(T_d\).
This process yields a sequence \((T_2,\dots,T_D)\) of structurally consistent warm starts with strong
objective values.

\begin{algorithm}[t]
\caption{\textsc{FeasibleSolutionInjection} for depth-\(D\) OCT (schematic)}
\label{alg:feasible-injection}
\begin{algorithmic}[1]
\Require \(X\in\mathbb{R}^{n\times p},y\); depth \(D\ge2\); RF ranking \(L\); feature increment \(k\)
\State Fit depth-\(D\) CART; \(\mathcal{F}_{\mathrm{CART}}\gets\) features used
\State Fit RF; obtain ranking \(L=(f^{(1)},\dots,f^{(p)})\)
\For{\(d=2,\dots,D\)}
    \If{\(d=2\)}
        \State \(\mathcal{F}_2 \gets \mathcal{F}_{\mathrm{CART}} \cup (\text{first }k\text{ in }L\setminus \mathcal{F}_{\mathrm{CART}})\)
        \State Solve OCT on \((X_{\mathcal{F}_2},y)\) \(\Rightarrow S^{\mathrm{warm}}_2\)
        \State Solve depth-\(2\) OCT on \((X,y)\) with \(S^{\mathrm{warm}}_2\) \(\Rightarrow T_2\)
    \Else
        \State \(\mathcal{F}_d \gets \mathcal{F}_{d-1} \cup (\text{next }k\text{ in }L\setminus \mathcal{F}_{d-1})\)
        \State Embed \(T_{d-1}\) into depth-\(d\) over \(\mathcal{F}_d\) \(\Rightarrow S^{\mathrm{sub}}_d\)
        \State Solve OCT on \((X_{\mathcal{F}_d},y)\) with \(S^{\mathrm{sub}}_d\) \(\Rightarrow \tilde{T}_d\)
        \State Lift \(\tilde{T}_d\) to full space \(\Rightarrow S^{\mathrm{full}}_d\)
        \State Solve depth-\(d\) OCT on \((X,y)\) with \(S^{\mathrm{full}}_d\) \(\Rightarrow T_d\)
    \EndIf
\EndFor
\State \Return \(T_D\)
\end{algorithmic}
\end{algorithm}

\subsubsection{Heuristic injection at MIP nodes}

Within each MIP solve, we further accelerate convergence by injecting additional feasible solutions at the nodes of a branch-and-bound tree.
The   procedure is detailed Algorithm \ref{alg:node-heuristic}.
Let \(\mathbf{b}^*=(b^*_{nf})\) be the LP-relaxation values of the branching variables at a node.
Define the saliency score
\[
\mathrm{score}(f) \;=\; \sum_{n\in\mathcal{B}} b^*_{nf}, \qquad f\in\mathcal{F},
\]
and let \(\mathcal{S}_{\mathrm{inc}}\subseteq\mathcal{F}\) be the feature set of the incumbent tree.
We form a candidate feature subset
\[
\mathcal{S}_1 \;=\; \mathcal{S}_{\mathrm{inc}}
\cup \Bigl\{\text{top-3 } f \in \mathcal{F}\setminus\mathcal{S}_{\mathrm{inc}} \text{ by } \mathrm{score}(f)\Bigr\}.
\]
A history set \(\mathcal{H}\) stores previously tried subsets; if \(\mathcal{S}_1\in\mathcal{H}\), the
attempt is skipped.
Heuristic injections are scheduled only at selected nodes.

On \(\mathcal{S}_1\), we solve a short-horizon sub-MIP to obtain an integer tree, lift it to a full
solution \((\mathbf{b},\mathbf{p},\mathbf{c},\mathbf{g})\), and inject it via
\texttt{cbSetSolution()} and \texttt{cbUseSolution()} whenever the objective improves the incumbent.

\begin{algorithm}[t]
\caption{\textsc{NodeHeuristicInjection} at a MIP node (schematic)}
\label{alg:node-heuristic}
\begin{algorithmic}[1]
\Require LP solution \(\mathbf{b}^*\); incumbent feature set \(\mathcal{S}_{\mathrm{inc}}\); history \(\mathcal{H}\)
\State \(\mathrm{score}(f) \gets \sum_{n\in\mathcal{B}} b^*_{nf}\) for all \(f\in\mathcal{F}\)
\State \(\mathcal{S}_1 \gets \mathcal{S}_{\mathrm{inc}} \cup \{\text{top-3 } f\notin\mathcal{S}_{\mathrm{inc}}\text{ by }\mathrm{score}(f)\}\)
\If{\(\mathcal{S}_1 \in \mathcal{H}\)} \State \Return \EndIf
\State \(\mathcal{H} \gets \mathcal{H} \cup \{\mathcal{S}_1\}\)
\State Solve sub-MIP on \(\mathcal{S}_1\) \(\Rightarrow\) integer tree
\State Lift to full solution \((\mathbf{b},\mathbf{p},\mathbf{c},\mathbf{g})\)
\If{objective improves incumbent}
    \State \texttt{cbSetSolution()}, \texttt{cbUseSolution()}
\EndIf
\end{algorithmic}
\end{algorithm}

\section{Evaluation}
\label{sec:experiments}
In this section, we evaluate various versions of our model
and conduct a comparative analysis with the current state-of-the-art approaches.
\subsection{Experimental Setup}

\paragraph{Datasets.}
We use 50 datasets compiled from prior work \citep{aghaei2025,demirovic2023}, covering a wide range of sizes and class balances: 24 small datasets (fewer than 1{,}000 samples), 12 medium datasets (1{,}000–5{,}000 samples), and 12 large datasets (more than 5{,}000 samples). Detailed statistics are reported in Table~\ref{uniquetsummary}. Following common practice, datasets containing categorical and/or continuous features are binarized via supervised discretization using the Minimum Description Length Principle (MDLP) \citep{fayyad1993}, followed by one-hot encoding. 
While alternative discretization strategies may improve downstream performance, a thorough investigation is left to future work.

\paragraph{Model training.}
For each dataset, we construct five independent random splits into training, validation,
and test sets with proportions 50\%/25\%/25\%, consistent with prior studies
\citep{aghaei2025}. 
All MIP models  are solved using Gurobi~11.0 \citep{optimization2023}.
The binarized datasets are publicly available at \url{https://github.com/Tommytutu/Benders-cut}, and
the source code will be released subsequently.

\subsection{Comparison to State-of-the-Art Methods}
In this section,  we evaluate the scalability of \texttt{BendersOCT-cut} relative to state-of-the-art open-source
baselines.
The first is the  MIP based formulation \texttt{BendersOCT}  \citep{aghaei2025} which has shown to be more scalable than the previous MIP based models.
The second is the DP based 
model, \texttt{BendersOCT} \citep{Subramanian2023} which is much more scalable than the existing MIP based models and can generalize to  nonlinear classification metrics such as F1-score.

\subsubsection{Comparing with BendersOCT}
We compare the \texttt{BendersOCT} baseline \citep{aghaei2025} with the following two  variants show the efficiency of the cutting planes and instance reduction method:
\begin{itemize}
    \item {\texttt{BendersOCT-weighted}}: the weighted master formulation in \eqref{eq:bendersOCT_master_2}.
    \item {\texttt{BendersOCT-weighted-cut}}: \texttt{BendersOCT-weighted} solved with the branch and cut algorithm described in Section~\ref{subsec:b&c}.
\end{itemize}

\begin{figure}[]
    \begin{minipage}[t]{0.5\linewidth}
        \centering
        \includegraphics[width=\textwidth]{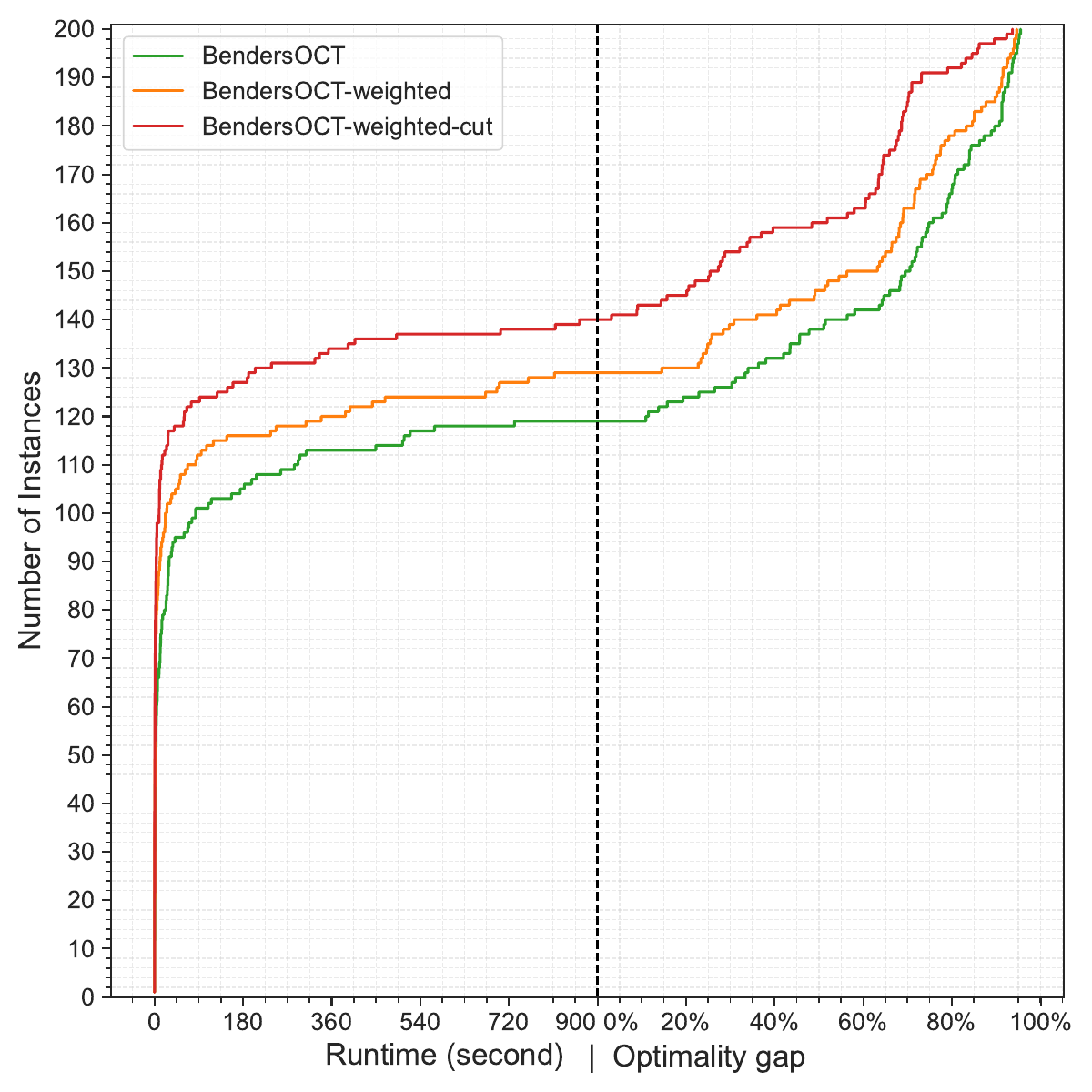}
        \centerline{(a) }
    \end{minipage}%
    \begin{minipage}[t]{0.5\linewidth}
        \centering
        \includegraphics[width=\textwidth]{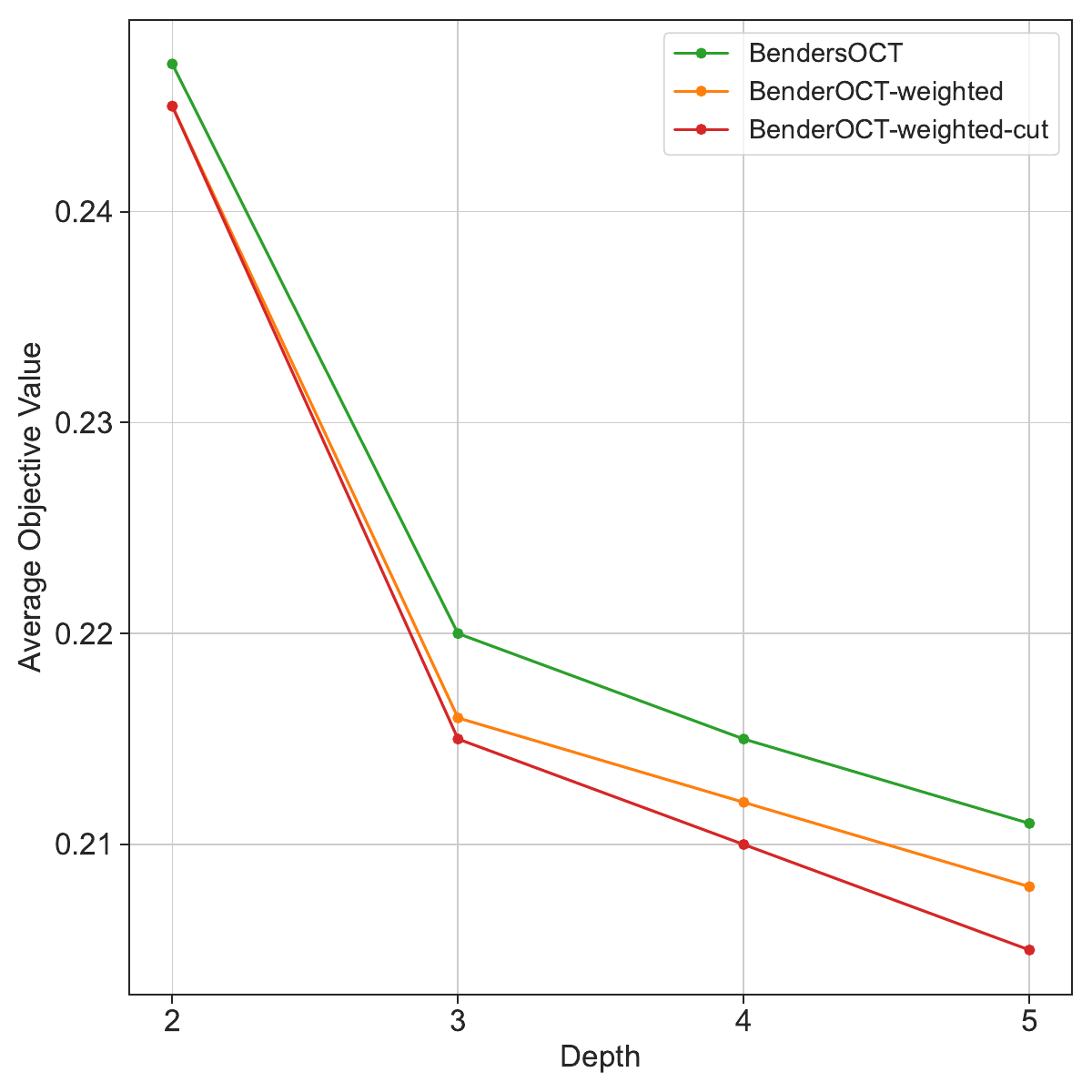}
        \centerline{(b)}
    \end{minipage}
    \caption{The results of MIP-based models displaying (a)  instances reaching optimality over time and
instances attaining a prescribed optimality-gap threshold at the time limit, (b)  the  average objective value  by each approach across depths $d\in$\{2,3,4,5\}.}
\label{fig:time_gap_obj}
\end{figure}

For fair comparison, we don't use any warm-start solutions for the MIP models (\texttt{BendersOCT}, \texttt{BendersOCT-weighted} and \texttt{BendersOCT-weighted-cut}).
Set \texttt{$\lambda =0.01$}  and \texttt{runtime = 900} seconds, we train the MIP models on the training set and report the runtime, objective value and optimality gap.
The optimality gap is defined as
$
\text{Gap} \;=\; 100\times\frac{\text{UB}-\text{LB}}{\text{UB}}\,\%,$
where UB is the best incumbent objective and LB is the global bound at termination.
Figure~\ref{fig:time_gap_obj} summarizes in-sample performance in terms of optimality gap, runtime, and objective values.

In Figure \ref{fig:time_gap_obj} (a),  along the time axis, \texttt{BendersOCT-weighted-cut} solves the largest number of instances within the time limit, followed by \texttt{BendersOCT-weighted}, with \texttt{BendersOCT} being the slowest across all depths. 
Specifically, \texttt{BendersOCT-weighted-cut} solves 140 instances to optimality within 900s, \texttt{BendersOCT-weighted} solves 129 instances to optimality within 900s, \texttt{BendersOCT} solves 118 instances to optimality within 900s.
\texttt{BendersOCT-weighted-cut} can solve 118 instances to optimality   with less than 45 seconds, yielding more than a $\frac{900}{45}=20\times$ speedup over \texttt{BendersOCT}. 
Along the gap axis, \texttt{BendersOCT-weighted-cut} consistently achieves smaller optimality gaps than both alternatives.
As  Figure \ref{fig:time_gap_obj} (b) shows, \texttt{BendersOCT-weighted-cut} can obtain better objective values as depth increases.

\subsubsection{Comparison with \texttt{StreeD}}
We compare our model with methods designed for optimizing nonlinear classification metrics.
\citet{Lin2020} proposed a dynamic programming (DP) approach for objectives such as F1-score, but observed that the F1-score is harder to optimize since optimal label assignments in different leaves are interdependent. They introduced a heuristic parameter~$\omega$ to simplify the problem, sacrificing optimality.
\citet{demirovic2021} developed a DP-based method capable of optimizing functions monotonic in both class misclassification scores (e.g., F1-score, Matthews correlation coefficient), but it lacks scalability—failing to find feasible solutions within one hour for many depth-4 instances (see Table~2 in their paper).
\citet{linden2023} improved scalability and proposed an anytime algorithm, \texttt{StreeD}. However, in their released implementation, only F1-score maximization is supported despite claims of generality to other nonlinear metrics.
Therefore, we compare our \texttt{BendersOCT-cut} with \texttt{StreeD} on the task of maximizing F1-score.
For a fair comparison, we set the model complexity parameter $\lambda$ to a small value of $0.0001$ for \texttt{BendersOCT-cut}. 
In addition, we fix the maximum number of branching nodes to $2^{D}-1$ (\texttt{max\_num\_nodes} $= 2^{D}-1$), corresponding to a full binary tree of depth $D$.

We use the 18 binary datasets on which \citet{demirovic2021} failed to find depth-4 solutions, setting a 90-second time limit.
Performance is evaluated by in-sample F1-score and runtime (Table~\ref{StreedvsBenders}, best results in bold).
At depth~4, \texttt{BendersOCT-cut} achieves the best F1-score on 6 datasets versus 3 for \texttt{StreeD}.
At depth~5, \texttt{BendersOCT-cut} demonstrates superior scalability, achieving the best results on 11 datasets, while \texttt{StreeD} performs best on only 2 and fails to find feasible solutions for 5 datasets.
Notably, on the \texttt{appendicitis} dataset, \texttt{BendersOCT-cut} attains the optimal F1-score (0.889), whereas \texttt{StreeD} produces a suboptimal result (0.75).

\begin{table}[ht]
\centering
\caption{The in-sample F1-score  and runtime (s) for maximizing F1-score by BendersOCT-cut and Streed. If no feasible solution is found within 90 seconds, then denote with `---'. Best results are shown in bold.}

\begin{threeparttable} 
\renewcommand{\arraystretch}{0.98}
\resizebox{0.96\textwidth}{!}{ 
\begin{tabular}{lcccccccc}
\toprule 
\multicolumn{1}{c}{\multirow{3}{*}{instance}} & \multicolumn{4}{c}{D=4}                                         & \multicolumn{4}{c}{D=5}                                    \\\cmidrule(lr){2-5}\cmidrule(lr){6-9}
\multicolumn{1}{c}{}                          & \multicolumn{2}{c}{F1-score}          & \multicolumn{2}{c}{Time}      & \multicolumn{2}{c}{F1-score}          & \multicolumn{2}{c}{Time} \\\cmidrule(lr){2-3}\cmidrule(lr){4-5}\cmidrule(lr){6-7}\cmidrule(lr){8-9}
\multicolumn{1}{c}{}                          & BendersOCT-cut     & Streed         & BendersOCT-cut    & Streed        & BendersOCT-cut     & Streed         & BendersOCT-cut     & Streed  \\\midrule
appendicitis                                  & \textbf{0.824} & 0.750          & \textbf{2.7}  & 90.2          & \textbf{0.889} & 0.750          & \textbf{4.7}   & 90.9    \\
biodeg                                        & 0.875          & 0.875 & 91.3          & 90.9          & 0.875          & \textbf{0.883} & 90.9           & 90.9    \\
cleve                                         & \textbf{0.931} & 0.866          & 91.8          & 90.9          & \textbf{0.937} & 0.866          & 90.3           & 90.7    \\
colic                                         & \textbf{0.959} & 0.947          & 90.4          & 90.9          & \textbf{0.959} & 0.947          & 90.5           & 90.8    \\
compas                                        & 0.689          & 0.689          & 91.2          & \textbf{28.7} & 0.689          & 0.689          & 90.2           & 90.5    \\
diabetes                                      & 0.869          & 0.869          & 90.5          & 90.1          & \textbf{0.898} & ---              & 91.3           & 90.6    \\
fico                                          & 0.721          & 0.721          & 91.0          & 90.9          & \textbf{0.730} & ---              & 93.0           & 91.0    \\
german                                        & 0.874          & 0.874          & 90.5          & 91.0          & \textbf{0.894} & 0.874          & 90.5           & 90.9    \\
heart                                         & \textbf{0.930} & 0.862          & 90.3          & 90.9          & \textbf{0.945} & 0.862          & 90.3           & 90.9    \\
hepatitis                                     & 1.000          & 1.000          & 96.7          & 90.9          & 1.000          & 1.000          & 91.1           & 90.7    \\
HTRU\_2                                       & 0.876          & \textbf{0.892} & 90.5          & 90.8          & \textbf{0.902}          & ---              & 90.5           & 90.4    \\
hungarian                                     & \textbf{0.844} & 0.835          & 91.3          & 90.9          & \textbf{0.844} & 0.835          & 94.4           & 90.9    \\
magic04                                       & \textbf{0.751} & 0.750          & 95.8          & 90.8          & 0.751          & 0.751          & 90.7           & 102.8   \\
pendigits                                     & 0.999          & 0.999          & 97.6          & 90.8          & 0.999          & \textbf{1.000} & 91.0           & 90.5    \\
promoters                                     & 1.000          & 1.000          & \textbf{11.8} & 52.5          & 1.000          & 1.000          & \textbf{14.2}  & 90.6    \\
spambase                                      & 0.889          & 0.889          & 95.4          & 90.3          & \textbf{0.889} & ---              & 90.4           & 90.7    \\
vehicle                                       & 0.917          & 0.917          & 91.5          & 90.9          & 0.917          & 0.917          & 91.3           & 90.3    \\
yeast                                         & 0.615          & 0.615          & 90.5          & 90.9          & \textbf{0.615} & ---              & 90.6           & 90.8   
\\ \bottomrule 
\end{tabular}
}
\end{threeparttable}
\label{StreedvsBenders}
\end{table}

\subsection{Ablation Study}
\subsubsection{Benefits  of Feasible Solution Injections}

Table~\ref{Abstudy_injection} compares the proposed feasible-solution injection strategy
(Section~\ref{subsec:feasible-injection}) with a CART-based warm start, averaged over 50 datasets
for depths \(d=2,\dots,5\).
\emph{Injection Objval} reports the objective value of the initial incumbent produced by each
warm-start strategy, while \emph{Final Objval} and \emph{Final Gap} correspond to the best
objective value and MIP gap after 900 seconds of Gurobi runtime.

Across all depths, our method yields consistently better initial incumbents than the CART warm
start (e.g., \(0.239\) vs.\ \(0.262\) at \(d=2\), \(0.199\) vs.\ \(0.224\) at \(d=5\)),
showing that the proposed warm-start method construct substantially stronger starting solutions.
These advantages persist and even amplify after full optimization: the final objective values 
under our warm start are uniformly lower than those under CART (e.g., \(0.229\) vs.\ \(0.245\) at
\(d=2\), \(0.195\) vs.\ \(0.202\) at \(d=5\)).
Moreover, the average final gap is reduced by roughly \(10\)–\(14\) percentage points across
depths (from \(18.3\%\) to \(4.0\%\) at \(d=2\), and from \(35.6\%\) to \(26.3\%\) at \(d=5\)),
indicating faster convergence and tighter bounds.
Overall, the proposed feasible-solution injection framework substantially strengthens both the
initial and final performance of the branch-and-cut procedure compared with a standard CART-based
warm start.

\begin{table}[ht]
\centering
\caption{The comparative results of the two warm start strategies. }

\begin{threeparttable}
\begin{tabular}{cccccccc}
\toprule  
      & \multicolumn{2}{c}{Injection Objval} & \multicolumn{2}{c}{Final Objval} & \multicolumn{2}{c}{Final Gap} \\\cmidrule(lr){2-3}\cmidrule(lr){4-5}\cmidrule(lr){6-7}
Depth & This paper       & CART       & This paper     & CART     & This paper    & CART     \\\midrule
2     & \textbf{0.239}            & 0.262      & \textbf{0.229}          & 0.245     & \textbf{4.0\% }       & 18.3\%   \\
3     & \textbf{0.208}            & 0.242      & \textbf{0.203}          & 0.215     & \textbf{13.8\%}       & 28.4\%   \\
4     & \textbf{0.201}            & 0.228      & \textbf{0.199}          & 0.211     & \textbf{22.3\%}       & 33.3\%   \\
5     & \textbf{0.199}            & 0.224      & \textbf{0.195}          & 0.202     & \textbf{26.3\% }      & 35.6\%  
\\ \bottomrule 
\end{tabular}
\end{threeparttable}
\label{Abstudy_injection}
\end{table}

\subsubsection{Benefits of Linearization}
\label{ref:experiment sub_Linearization} 
In this section, we experiment with the two kinds of formulations (MIP and MIQP)  of the optimal classification trees with maximizing different objectives.
For limited space, we only report the results of maximizing F1-score and  MCC  with depth=5 as shown in Table \ref{table:insample_F1}, and the results on the other nonlinear objectives are similar to  MCC.

For maximizing F1-score,  both the MIP and MIQP formulations  can solve  6 instances (out of 18 instances) to optimality within the timelimit (900 seconds) at the depth of 5, while the method proposed by \cite{demirovic2021} cannot find a feasible solution within  1 hour for these   datasets (refer to the Table 2 in \cite{demirovic2021}.
When compared with the two formulations for maximizing F1-score, the MIP formulation performs better than MIQP in terms of gap, time and objective value.  
This shows that the  linearization can improve  scalability.
Therefore, we can infer that our MIP formulation for the F1-score is more scalable than the MIQP proposed by \cite{tu2025}. Moreover, the same linearization techniques can also be applied to their MIQP model.
For maximizing MCC, MIP formulation performs better than MIQP in terms of gap, time and objective value.  

\begin{table}[ht]
\centering
\caption{Results of the Two Formulations for  Maximizing F1-score and MCC with Depth =5.}
\begin{threeparttable} 
\renewcommand{\arraystretch}{0.98}
\resizebox{0.96\textwidth}{!}{ 
\begin{tabular}{llllllllllllll}
\toprule 
\multicolumn{1}{c}{\multirow{3}{*}{Instance}} & \multicolumn{6}{c}{F1-score}                                                                            & \multicolumn{6}{c}{MCC}                                                                               \\\cmidrule(lr){2-7}\cmidrule(lr){8-13}
\multicolumn{1}{c}{}                          & \multicolumn{2}{c}{Gap}           & \multicolumn{2}{c}{Time}       & \multicolumn{2}{c}{ObjVal} & \multicolumn{2}{c}{Gap}           & \multicolumn{2}{c}{Time}     & \multicolumn{2}{c}{ObjVal}      \\\cmidrule(lr){2-3}\cmidrule(lr){4-5}\cmidrule(lr){6-7}\cmidrule(lr){8-9}\cmidrule(lr){10-11}\cmidrule(lr){12-13}
\multicolumn{1}{c}{}                          & MIP             & MIQP            & MIP             & MIQP            & MIP     & MIQP             & MIP             & MIQP            & MIP             & MIQP          & MIP            & MIQP           \\\midrule
appendicitis                                  & 0                & 0                & \textbf{0.04}   & 0.05          & 0.04           & 0.04           & 0.0\%           & 0.0\%   & \textbf{0.3}     & 33.7     & 0.396            & 0.396   \\
biodeg                                        & \textbf{74.10\%} & 77.30\%          & 900.5           & 901.3         & \textbf{0.289} & 0.299          & \textbf{88.1\%} & 100.0\% & 903.2            & 900.3    & 0.554            & 0.554   \\
cleve                                         & 0                & 0                & \textbf{183.65} & 316.07        & 0.092          & 0.092          & \textbf{75.3\%} & 96.9\%  & 901.6            & 900.4    & \textbf{0.397}   & 0.408   \\
colic                                         & \textbf{11.20\%} & 12.60\%          & 900.22          & 900.13        & 0.103          & 0.103          & \textbf{24.3\%} & 71.5\%  & 900.1            & 900.4    & \textbf{0.227}   & 0.365   \\
compas                                        & 0                & 0                & 0.77            & \textbf{0.49} & 0.367          & 0.367          & \textbf{0.0\%}  & 100.0\% & \textbf{6.5}     & 900.3    & \textbf{0.922}   & 0.957   \\
diabetes                                      & 64.70\%          & \textbf{61.90\%} & 901.16          & 900.68        & 0.176          & 0.176          & \textbf{96.1\%} & 100.0\% & 903.1            & 900.3    & \textbf{0.787}   & 0.814   \\
fico                                          & \textbf{37.50\%} & 47.90\%          & 901.22          & 901.02        & \textbf{0.317} & 0.353          & \textbf{75.8\%} & 100.0\% & 901.8            & 900.5    & 0.911            & 0.911   \\
german                                        & \textbf{55.90\%} & 57.50\%          & 900.53          & 900.38        & 0.176          & 0.176          & \textbf{96.6\%} & 100.0\% & 901.2            & 900.4    & \textbf{0.879}   & 0.909   \\
heart                                         & \textbf{38.40\%} & 50.20\%          & 900.11          & 900.52        & \textbf{0.151} & 0.161          & \textbf{70.4\%} & 97.4\%  & 901.5            & 900.3    & \textbf{0.420}   & 0.518   \\
hepatitis                                     & 0                & 0                & \textbf{1.56}   & 3.74          & 0.06           & 0.06           & 0.0\%           & 0.0\%   & \textbf{10.1}    & 43.9     & 0.100            & 0.100   \\
HTRU\_2                                       & \textbf{23.90\%} & 28.20\%          & 900.64          & 900.73        & \textbf{0.169} & 0.203          & \textbf{0.0\%}  & 100.0\% & \textbf{2.8}     & 900.7    & \textbf{0.304}   & 0.503   \\
hungarian                                     & \textbf{40.90\%} & 42.00\%          & 900.19          & 900.14        & 0.127          & 0.127          & \textbf{59.0\%} & 93.7\%  & 900.8            & 900.2    & \textbf{0.428}   & 0.571   \\
magic04                                       & \textbf{80.60\%} & 80.90\%          & 901.1           & 900.64        & \textbf{0.355} & 0.48           & \textbf{87.3\%} & 91.9\%  & 900.9            & 900.5    & \textbf{0.817}   & 0.924   \\
pendigits                                     & 0           & {0}       & \textbf{556.35}          & 769.29        & 0.076          & 0.076          & \textbf{55.1\%} & 91.9\%  & 901.7            & 900.5    & \textbf{0.114}   & 0.124   \\
promoters                                     & 0                & 0                & \textbf{0.14}   & 0.17          & 0.04           & 0.04           & 0.0\%           & 0.0\%   & \textbf{1.0}     & 4.8      & 0.070            & 0.070   \\
spambase                                      & 88.20\%          & \textbf{82.60\%} & 903.76          & 900.88        & 0.185          & \textbf{0.179} & \textbf{95.2\%} & 100.0\% & 901.1            & 900.7    & \textbf{0.434}   & 0.460   \\
vehicle                                       & \textbf{32.10\%} & 38.30\%          & 900.29          & 900.28        & 0.116          & 0.116          & \textbf{74.6\%} & 99.9\%  & 901.4            & 900.0    & \textbf{0.234}   & 0.292   \\
yeast                                         & \textbf{89.50\%} & 92.00\%          & 901             & 29116.63      & \textbf{0.431} & 0.453          & \textbf{96.6\%} & 100.0\% & 901.8            & 900.5    & 0.894            & 0.894  \\
average & \textbf{35.82\%} & 37.30\% & 660.98 & 660.95 & \textbf{0.182} & 0.195 & \textbf{55.3\%} & 80.2\% & \textbf{652.3} & 754.9 & \textbf{0.494} & 0.543 
\\ \bottomrule         
\end{tabular}
}
\end{threeparttable}
\label{table:insample_F1}
\end{table}

\subsubsection{Benefits  on Nonlinear Objectives}
We examine whether directly optimizing nonlinear objectives (e.g., F1-score and MCC) yields better generalization on imbalanced datasets compared with accuracy maximization.  
Table~\ref{table:outsampleF1MCC} reports out-of-sample results for depth-5 optimal trees trained under three objectives: accuracy (\(\text{obj}=\mathrm{Acc}\)), F1-score (\(\text{obj}=\mathrm{F1}\)), and MCC (\(\text{obj}=\mathrm{MCC}\)).  
For each dataset and evaluation metric (Accuracy, F1-score, MCC), the best result is highlighted in bold.

As shown in Table~\ref{table:outsampleF1MCC}, models trained to maximize F1-score or MCC generalize well to the test sets.  
In terms of test \emph{accuracy}, the MCC-based model performs best on 11 of 18 datasets (including ties), compared with 7 each for the accuracy- and F1-based models.  
For test \emph{F1-score}, the F1- and MCC-based models achieve 8 and 9 best results, respectively, both outperforming the accuracy-based model on several imbalanced datasets (e.g., \texttt{compas}, \texttt{yeast}). For test \emph{MCC}, the MCC-based model leads on 11 datasets, clearly dominating the accuracy-based model.

Averaged results confirm these trends.  
Optimizing F1-score or MCC increases the mean test F1 from \(0.717\) (accuracy objective) to \(0.760\) and \(0.753\), representing absolute gains of \(4.3\%\) and \(3.6\%\).  
Optimizing MCC raises the mean test MCC from \(0.551\) to \(0.575\) (a \(2.4\%\) improvement) while slightly improving accuracy (\(0.808 \rightarrow 0.811\)).  
Overall, optimizing nonlinear objectives enhances nonlinear performance metrics on imbalanced test sets while maintaining or modestly improving accuracy relative to accuracy-only training.

\begin{table}[ht]
\centering
\caption{Out-of-sample results of different objectives when depth =5.}
\begin{threeparttable}  
\renewcommand{\arraystretch}{0.95}
\resizebox{0.95\textwidth}{!}{ 
\begin{tabular}{lcccccccccc}
\toprule 
             & \multicolumn{3}{c}{Acuracy}                      & \multicolumn{3}{c}{F1-score}                       & \multicolumn{3}{c}{MCC}                      \\\cmidrule(lr){2-4}\cmidrule(lr){5-7}\cmidrule(lr){8-10}
instance     & obj=Acc      & obj=F1             & obj=MCC       &  obj=Acc       & obj=F1             & obj=MCC              &  obj=Acc &obj=F1             & obj=MCC            \\\midrule
appendicitis & \textbf{1.000}          &\textbf{1.000}          & \textbf{1.000}          & \textbf{1.000}         & \textbf{1.000}          & \textbf{1.000}          & \textbf{1.000}        & \textbf{1.000}          & \textbf{1.000}         \\
biodeg       & 0.795          & 0.795          & \textbf{0.822} & 0.679          & 0.679          & \textbf{0.734} & 0.529          & 0.529          & \textbf{0.601} \\
cleve        & 0.789          & \textbf{0.816} & 0.789          & 0.714          & \textbf{0.788} & 0.714          & 0.567          & \textbf{0.626} & 0.567          \\
colic        & 0.793          & \textbf{0.815} & 0.783          & 0.843          & \textbf{0.857} & 0.831          & 0.579          & \textbf{0.623} & 0.549          \\
compas       & 0.658          & 0.594          & \textbf{0.672} & 0.581          & \textbf{0.651} & 0.617          & 0.303          & 0.254          & \textbf{0.333} \\
diabetes     & 0.786          & 0.781          & \textbf{0.802} & \textbf{0.861} & 0.850          & 0.852          & 0.501          & 0.475          & \textbf{0.555} \\
fico         & \textbf{0.700} & 0.698          & 0.698          & 0.641          & \textbf{0.681} & \textbf{0.681} & \textbf{0.402} & 0.277          & 0.277          \\
german       & 0.740          & 0.740          & 0.684          & 0.846          & \textbf{0.851} & 0.683          & \textbf{0.123} & 0.000          & 0.114          \\
heart        & \textbf{0.882} & 0.809          & 0.868          & \textbf{0.826} & 0.745          & 0.809          & 0.745          & 0.624          & \textbf{0.719} \\
hepatitis    & \textbf{0.718} & \textbf{0.718} & \textbf{0.718} & 0.353          & 0.353          & 0.353          & 0.299          & 0.299          & 0.299          \\
HTRU2        & 0.973          & \textbf{0.979} & \textbf{0.979} & 0.828          & \textbf{0.871} & \textbf{0.871} & 0.822          & \textbf{0.862} & \textbf{0.862} \\
hungarian    & 0.770          & 0.784          & \textbf{0.797} & 0.638          & 0.692          & \textbf{0.706} & 0.531          & 0.547          & \textbf{0.579} \\
magic04      & \textbf{0.727} & \textbf{0.727} & \textbf{0.727} & 0.666          & 0.666          & 0.666          & 0.457          & 0.457          & 0.457          \\
pendigits    & 0.986          & 0.990          & \textbf{0.993} & 0.924          & 0.948          & \textbf{0.965} & 0.917          & 0.943          & \textbf{0.961} \\
promoters    & \textbf{0.741} & \textbf{0.741} & \textbf{0.741} & 0.759          & 0.759          & 0.759          & 0.532          & 0.532          & 0.532          \\
spambase     & 0.855          & 0.855          & \textbf{0.862} & 0.815          & 0.815          & \textbf{0.826} & 0.697          & 0.697          & \textbf{0.712} \\
vehicle      & 0.920          & 0.953          & \textbf{0.958} & 0.847          & 0.911          & \textbf{0.920} & 0.793          & 0.879          & \textbf{0.892} \\
yeast        & \textbf{0.709} & 0.704          & 0.704          & 0.085          & \textbf{0.563} & \textbf{0.563} & 0.126          & \textbf{0.349} & \textbf{0.349}  \\\midrule  
average      & 0.808          & 0.805          & \textbf{0.811}          & 0.717          & \textbf{0.760}          & 0.753          & 0.551          & 0.554          & \textbf{0.575}      
\\ \bottomrule         
\end{tabular}
}
\end{threeparttable}
\label{table:outsampleF1MCC}
\end{table}

\section{Conclusion}
\label{sec:conclusion}

We developed a general framework for learning optimal classification trees via mixed-integer programming. 
The approach (i) compresses the training data through an instance-reduction scheme that aggregates duplicate instances with weights; 
(ii) builds on the Benders decomposition of \citet{aghaei2025} and strengthens the master problem using conflict-subset and feature-activated cutting planes; and 
(iii) extends  to nonlinear, imbalance-aware objectives (e.g., $F_\beta$ and MCC) through exact mixed-integer linear reformulations. 
Empirically, the proposed acceleration mechanisms—including a tailored branch-and-cut algorithm, instance reduction, and warm-start strategies—yield substantial improvements in scalability relative to the state-of-the-art MIP formulation \texttt{BendersOCT}. 
For nonlinear objectives, the resulting MIP models scale markedly better than MIQP counterparts and the DP-based approach \texttt{StreeD} \citep{linden2023}, and trees trained to directly optimize nonlinear metrics often generalize better on those metrics for imbalanced data.

Despite the substantial gains, several challenges remain. First, optimizing nonlinear objectives (e.g., F1-score and MCC) at larger depths can still be computationally demanding due to the interaction of multiple count products and data-dependent binary variables. 
Second, performance depends on the quality of binarization; supervised discretization is convenient but may not be optimal for a given metric. 
Third, while our unique-instance reduction controls instance count, feature dimensionality still drives problem size.
Future work will focus on more scalable linearization techniques and stronger convex-hull formulations for nonlinear objectives. 
In addition, given the strong dependence of scalability on the number of features, we plan to explore effective feature-selection strategies specifically tailored to our formulation.

\begingroup
\setlength{\bibsep}{0pt}
\bibliographystyle{informs2014}
\bibliography{fodt}
\endgroup

\newpage

\ECSwitch

\ECHead{E-Companion}

\section{FlowOCT}
\label{appendix_sec: FlowOCT}

In this section, we provide a simplified version of the formulation of

\begin{subequations}
\begin{align}
\text{maximize} \;\; & \displaystyle (1-\lambda) \sum_{i \in \mathcal I} \sum_{n \in \mathcal{B} \cup  \mathcal{L} } z^i_{n,t}  - \lambda \sum_{n \in  \mathcal{V}}\sum_{f\in  \mathcal{F}}b_{nf} \label{eq:flow_reg_obj1}\\
\text{subject to}\;\;
& \displaystyle \sum_{f \in \mathcal{F}}b_{nf} + p_n + \sum_{m \in \mathcal{P}{(n)}}p_m = 1   &\hspace{-5cm}  \forall n \in \mathcal{B} \label{eq:flow_reg_branch_or_predict1}\\
&  p_n+\sum_{m \in  \mathcal{P}(n)}p_{m} =1   &  \forall n \in  \mathcal{T} \label{eq:flow_reg_terminal_leaf1}\\
& \displaystyle z^i_{a(n),n} =  z^i_{n,\ell(n)} + z^i_{n,r(n)} + z^i_{n,t}  &\hspace{-5cm}  \forall n \in \mathcal{B}, i \in \mathcal I \label{eq:flow_reg_conservation_internal1}\\
&  \displaystyle z^i_{a(n),n} = z^i_{n,t} &\hspace{-5cm}   \forall n \in \mathcal B \cup \mathcal L, {i \in \mathcal I} \label{eq:flow_reg_conservation_terminal1}\\
& \displaystyle z^i_{s,1} \leq 1 &\hspace{-5cm} \forall i \in \mathcal I\label{eq:flow_reg_source1}\\
&  \displaystyle z^i_{n,\ell(n)}\leq \sum_{f \in \mathcal F: x_{f}^i=0}b_{nf} &\hspace{-5cm} \forall n \in \mathcal B, i \in \mathcal I \label{eq:flow_reg_branch_left1}\\
&  \displaystyle z^i_{n,r(n)}\leq \sum_{f \in \mathcal F: x_{f}^i=1}b_{nf}  &\hspace{-5cm} \forall n \in \mathcal B, i \in \mathcal I \label{eq:flow_reg_branch_right1}\\
&  \displaystyle z^i_{n,t} \leq  c_{ny^i} &\hspace{-5cm} \forall  n \in \mathcal B \cup \mathcal L, {i \in \mathcal I} \label{eq:flow_reg_sink1}\\
&  \displaystyle \sum_{k \in \mathcal K}c_{nk} = p_n  &\hspace{-5cm}  \forall n \in \mathcal B \cup \mathcal T \label{eq:flow_reg_leaf_prediction1}\\
&  \displaystyle c_{nk} \in \{0,1\}  &\hspace{-5cm}   \forall n \in \mathcal B \cup \mathcal L,k \in \mathcal K \\
&  \displaystyle b_{nf} \in \{0,1\}  &\hspace{-5cm}   \forall n \in \mathcal B,f \in \mathcal F \\
&  \displaystyle p_{n} \in \{0,1\}  &\hspace{-5cm}   \forall n \in \mathcal B \cup \mathcal L \\
&  \displaystyle z^i_{a(n),n}, z^i_{n,t}\in \{0,1\}  &\hspace{-5cm}  \forall n \in \mathcal B \cup \mathcal L,i \in \mathcal I,
\end{align}
\label{eq:flow_reg}%
\end{subequations}
where  $\lambda\in [0,1]$ is a regularization parameter. An explanation of the  constraints is as follows. Constraint{s}~\eqref{eq:flow_reg_branch_or_predict1} {imply} that at any node $n \in \mathcal B$  we either branch on a feature $f$ (if~$\sum_{f \in \mathcal F}b_{nf}=1$), predict a label (if~$p_n=1$), or get pruned if a prediction is made at one of the node ancestors (i.e., if~$\sum_{m \in \mathcal P(n)}p_m=1$). Similarly constraint{s}~\eqref{eq:flow_reg_terminal_leaf1} {ensure} that any node $n \in \mathcal B \cup \mathcal L$ is either a leaf node of the tree or is pruned. 
Constraint{s}~\eqref{eq:flow_reg_conservation_internal1} {are} flow conservation constraints for each datapoint $i$ and node $n \in \mathcal B$: {they} ensure that if a datapoint arrives at a node, {then} it must also leave the node through one of its descendants. Similarly, constraint{s}~\eqref{eq:flow_reg_conservation_terminal1} enforce flow conservation for each node $n \in \mathcal L$.
The inequality constraint{s}~\eqref{eq:flow_reg_source1} imply that at most one unit of flow can enter the graph through the source {for each datapoint}. 
Constraint{s}~\eqref{eq:flow_reg_branch_left1} (resp.\ \eqref{eq:flow_reg_branch_right1}) ensure that if {the flow of }a datapoint is routed to the left (resp.\ right) at node~$n$, then one of the features such that $x^i_f=0$ (resp.\ $x^i_f=1$) must have been selected for branching at the node. 
Constraint{s}~\eqref{eq:flow_reg_sink1} guarantee that datapoints {whose flow is} routed to the sink node~$t$ are correctly classified.
Constraint{s}~\eqref{eq:flow_reg_leaf_prediction1} make sure that each leaf node is assigned a predicted class $k \in \mathcal K$. The objective \eqref{eq:flow_reg_obj1} maximizes the total number of correctly classified datapoints.

\section{Proof}
\subsection{Proof of Proposition \ref{Proposition:WFlowOCTeFlowOCT}}
Map each unique instance $i\in\mathcal{U}$ to the $w_i$ duplicates in $\mathcal{I}$. Any feasible \texttt{FlowOCT} solution $(b,p,c,z)$ on $\mathcal{I}$ induces a feasible \texttt{WFlowOCT} solution by setting the same $(b,p,c)$ and defining $z^i_{\cdot,\cdot}$ equal to the (common) flow pattern taken by its duplicates; then $\sum_{n} w_i z^i_{n,t}$ equals the total correct flow over all duplicates, so the objective values match (the regularization term is identical). Conversely, replicate each unique instance $i$ exactly $w_i$ times to expand $(\mathcal{U},w)$ back to $\mathcal{I}$; any feasible \texttt{WFlowOCT} solution lifts to a feasible \texttt{FlowOCT} solution with the same $(b,p,c)$ and duplicated flows, again preserving the objective. Optimality is preserved in both directions.

\subsection{Proof of Proposition \ref{Proposition:eq:bendersOCT_master_2}}
Fix an incumbent solution $(\bar{\bm b},\bar{\bm c},\bar{\bm p},\bar{\bm g})$ of \eqref{eq:bendersOCT_master_2}. 
For any instance $i\in\mathcal{U}$, consider the instance-specific capacity graph $G^i(\bar{\bm b},\bar{\bm c})$ on $\mathcal V=\{s\}\cup\mathcal{B}\cup\mathcal{L}\cup\{t\}$ with arc capacities
\[
\mathrm{cap}(s,1)=1,\quad
\mathrm{cap}(n,\ell(n))=\sum_{f:\,x_f^i=0}\bar b_{nf},\quad
\mathrm{cap}(n,r(n))=\sum_{f:\,x_f^i=1}\bar b_{nf},\quad
\mathrm{cap}(n,t)=\bar c_{n,y_i},
\]
and zero elsewhere. Let 
\[
\kappa_i \;=\; \min_{\mathcal S\subseteq \mathcal V\setminus\{t\}:~s\in\mathcal S}\;
\sum_{(n_1,n_2)\in \mathcal C(\mathcal S)} \mathrm{cap}(n_1,n_2)
\quad\text{and}\quad 
\mathcal S_i^\star\in\arg\min(\cdot)
\]
be the $s$--$t$ min-cut value and an associated minimizer. Then the most violated inequality among the Benders cuts 
\eqref{eq:bendersOCT_master_2_benders_cut} for instance $i$ is 
\[
g^i \;\le\; \sum_{(n_1,n_2)\in \mathcal C(\mathcal S_i^\star)} c^i_{n_1,n_2}(\bm b,\bm c),
\]
and a violation occurs if and only if $\bar g^i>\kappa_i$. Consequently, separation of 
\eqref{eq:bendersOCT_master_2_benders_cut} reduces to computing an $s$--$t$ min-cut on $G^i(\bar{\bm b},\bar{\bm c})$ (as in Algorithm~2 of \citet{aghaei2025}), which can be done in polynomial time.

For fixed $(\bar{\bm b},\bar{\bm c},\bar{\bm p})$, the sub problem  for instance $i$ is a maximum $s$--$t$ flow on $G^i(\bar{\bm b},\bar{\bm c})$. 
By the max-flow/min-cut theorem, every $s$--$t$ cut $\mathcal S$ yields a valid upper bound 
$g^i \le \sum_{(n_1,n_2)\in \mathcal C(\mathcal S)} \mathrm{cap}(n_1,n_2)$, 
and the tightest bound equals the min-cut value $\kappa_i$. 
Therefore, the most violated inequality for $i$ is obtained from $\mathcal S_i^\star$; 
it is violated exactly when $\bar g^i>\kappa_i$. 
Standard max-flow algorithms (e.g., push–relabel or Dinic) compute $\kappa_i$ and a minimizing cut in polynomial time, which provides a strong separation oracle for \eqref{eq:bendersOCT_master_2_benders_cut}.

\section{Separation procedure}
\begin{algorithm}[h]
	\OneAndAHalfSpacedXI
		\caption{Separation procedure for constraints \eqref{eq:bendersOCT_master_2_benders_cut1}}
		\label{alg:cut_reg}
		 \textbf{Input:} $(\bm b,\bm c, \bm p,\bm g)\in \{0,1\}^{\mathcal B\cdot \mathcal F}\cdot \{0,1\}^{\mathcal V\cdot \mathcal K}\cdot {\{0,1\}^{\mathcal V}}\cdot \R^{\mathcal U} \text{  satisfying~\eqref{eq:flow_reg_branch_or_predict}, \eqref{eq:flow_reg_terminal_leaf}, \eqref{eq:flow_reg_leaf_prediction}; } \newline
		 \hspace*{\algorithmicindent} i~\in~\mathcal U:  \text{ datapoint used to generate the cut.}$ \newline
		 \textbf{Output:} $-1$ if all constraints \eqref{eq:bendersOCT_master_2_benders_cut1} corresponding to $i$ are satisfied;  \newline 
		 \hspace*{\algorithmicindent} source set $\sets S$ of min-cut otherwise.
	    \begin{algorithmic}[1]
 		\State \textbf{if }$g^i=0$ \textbf{ then return } $-1$\label{line:simple_reg}
 		\State \textbf{Initialize} $n\leftarrow 1$ \hfill \Comment{Current node $=$ root}
 		\State \textbf{Initialize} $\sets S\leftarrow \{s\}$ \hfill \Comment{$\sets S$ is in the source set of the cut}
 		\While{{$p_n=0$}}\label{line:ini_reg}
 		\State $\sets S\leftarrow \sets S\cup\{n\}$
 		\If{$c_{n,\ell(n)}^i({\bm b},{\bm w})=1$}  \label{lin:sub-start_reg}
 		\State $n\leftarrow \ell(n)$ \hfill \Comment{Datapoint $i$ is routed left}
 		\ElsIf{$c_{n,r(n)}^i({\bm b},{\bm c})=1$} \label{lin:sub-start2_reg}
 		\State $n\leftarrow r(n)$ \hfill \Comment{Datapoint $i$ is routed right} \label{lin:sub-end_reg}
 		\EndIf
 		\EndWhile \label{line:end_reg}\Comment{\underline{At this point, $n$ is a leaf node of the tree}}
 		\State $\sets S\leftarrow \sets S\cup\{n\}$ \label{line:terminal_start_reg}
 		\If{$g^i > c_{n,t}^i({\bm b},{\bm c})$} \Comment{Minimum cut $\sets S$ with capacity 0 found}\label{line:separation_reg}
 		\State \textbf{return }$\sets S \label{line:return_reg}$
 		\Else \Comment{Minimum cut $\sets S$ has capacity 1, constraints \eqref{eq:bendersOCT_master_2_benders_cut1} are satisfied}\label{line:satisfaction_reg}
 		\State \textbf{return} $-1 \label{line:-1_reg}$
 		\EndIf\label{line:terminal_end_reg}
	\end{algorithmic}
\label{alg:Separation}
\end{algorithm}

\subsection{Proof of Proposition \ref{Proposition:Apositive}}
The sign of \(\mathrm{MCC}\) depends on \(A\).
We show that there always exists a feasible solution with \(A\ge 0\), i.e., \(\mathrm{MCC}\ge 0\).
If we classify all examples as negative, then \(\mathrm{TP}=0\) and \(\mathrm{TN}=n^-\), so
\[
A = n^+ n^- + n^- \cdot 0 - n^+ n^- = 0.
\]
Thus there exists a feasible solution with \(\mathrm{MCC}=0\), and maximizing \(\mathrm{MCC}\) is equivalent to maximizing \(\mathrm{MCC}^2\). \hfill$\square$

\subsection{Proof of Theorem \ref{thm:conflict_subset_improvement}}
No single leaf can correctly assign different labels to instances with identical $\mathbf{x}$. At most one label $k$ can be correctly assigned for all instances in $\mathcal{G}_s$, hence the bound \eqref{equation:lowers}.

\subsection{Proof of Proposition \ref{prop:Fs}}
For $i\in\mathcal{G}_s'$, let $\mathbf{x}_i|_{\mathcal{F}\setminus\mathcal{F}'}$ denote the projection of $\mathbf{x}_i$ onto the coordinates in $\mathcal{F}\setminus\mathcal{F}'$. By assumption, removing $\mathcal{F}'$ merges the involved unique instances into a conflict subset, hence
\[
\mathbf{x}_i|_{\mathcal{F}\setminus\mathcal{F}'} \;=\; \mathbf{x}_j|_{\mathcal{F}\setminus\mathcal{F}'}
\quad \text{for all } i,j\in\mathcal{G}_s',
\]
while their labels are not all identical. We analyze two cases.

\smallskip
\noindent\emph{Case 1:} $\displaystyle \sum_{n\in\mathcal{B}}\sum_{f\in\mathcal{F}'} b_{nf}=0$.  
No split in the tree uses a feature from $\mathcal{F}'$, so the routing of any sample depends only on $\mathbf{x}|_{\mathcal{F}\setminus\mathcal{F}'}$. Because all items in $\mathcal{G}_s'$ have identical projections on these features, they follow the \emph{same} path and reach the \emph{same} leaf $\ell$. Leaf $\ell$ predicts a single class $\hat{k}\in\mathcal{K}$; therefore, among the items in $\mathcal{G}_s'$ the number correctly classified equals $|\mathcal{G}'_{s\hat{k}}|$, and
\[
\sum_{i\in\mathcal{G}_s'} g^i
\;=\; |\mathcal{G}'_{s\hat{k}}|
\;\le\; \max_{k\in\mathcal{K}} |\mathcal{G}'_{sk}|.
\]
This is precisely the (tight) conflict bound that holds when the distinguishing features are unavailable.

\noindent\emph{Case 2:} $\displaystyle \sum_{n\in\mathcal{B}}\sum_{f\in\mathcal{F}'} b_{nf}\ge 1$.  
In this case the right-hand side of \eqref{eq:Fs} satisfies
\[
\max_{k} |\mathcal{G}'_{sk}|
\;+\;
\Bigl(|\mathcal{G}_s'|-\max_{k}|\mathcal{G}'_{sk}|\Bigr)
\sum_{n,f\in\mathcal{F}'} b_{nf}
\;\ge\;
\max_{k} |\mathcal{G}'_{sk}|
\;+\;
\Bigl(|\mathcal{G}_s'|-\max_{k}|\mathcal{G}'_{sk}|\Bigr)\cdot 1
\;=\; |\mathcal{G}_s'|.
\]
Since $g^i\in\{0,1\}$, we always have $\sum_{i\in\mathcal{G}_s'} g^i \le |\mathcal{G}_s'|$. Hence the inequality \eqref{eq:Fs} holds trivially in this case.

Combining the two cases proves \eqref{eq:Fs} for all feasible $(g^i,b_{nf})$.

\begin{table}[ht]
\centering
\caption{The number of unique samples for the 50 datasets.}

\begin{threeparttable} 
\renewcommand{\arraystretch}{0.8} 
{
\begin{tabular}{lllllllllllll}
\toprule 
dataset         & $|\mathcal{F}|$  & $|\mathcal{K}|$ & $|\mathcal{I}|$ & $|\mathcal{U}|$ & dataset         &  & $|\mathcal{F}|$  & $|\mathcal{K}|$ & $|\mathcal{I}|$ &$|\mathcal{U}|$ \\\midrule
soybean         & 45       & 4      & 47      & 47                & tic             & 958     & 27       & 2  & 958     & 958                          \\
appendicitis    & 530      & 2      & 106     & 106               & german-credit   & 1000    & 112      & 2  & 1000    & 998                          \\
promoters       & 334      & 2      & 106     & 106               & MaternalHealth  & 1014    & 23       & 3  & 1014    & 163                          \\
monk3           & 15       & 2      & 122     & 122               & biodeg          & 1055    & 81       & 2  & 1055    & 931                          \\
monk1           & 15       & 2      & 124     & 124               & messidor        & 1151    & 24       & 2  & 1151    & 231                          \\
hayes           & 15       & 3      & 132     & 78                & banknote        & 1372    & 16       & 2  & 1372    & 52                           \\
iris            & 12       & 3      & 150     & 24                & contraceptive   & 1473    & 21       & 3  & 1473    & 531                          \\
hepatitis       & 361      & 2      & 155     & 155               & yeast           & 1484    & 89       & 2  & 1484    & 1423                         \\
monk2           & 15       & 2      & 169     & 169               & car             & 1728    & 19       & 4  & 1728    & 1327                         \\
wine            & 32       & 3      & 178     & 128               & wireless        & 2000    & 42       & 4  & 2000    & 1199                         \\
spect           & 22       & 2      & 267     & 228               & kr              & 3196    & 65       & 2  & 3196    & 3173                         \\
heart           & 381      & 2      & 270     & 270               & students        & 4424    & 96       & 3  & 4424    & 4267                         \\
breast          & 36       & 2      & 277     & 263               & spambase        & 4601    & 132      & 2  & 4601    & 3594                         \\
hungarian       & 330      & 2      & 294     & 293               & compas          & 6907    & 12       & 2  & 6907    & 213                          \\
cleve           & 395      & 2      & 303     & 302               & pendigits       & 7494    & 216      & 2  & 7494    & 7415                         \\
column\_3c      & 15       & 3      & 310     & 80                & avila           & 10430   & 128      & 12 & 10430   & 8017                         \\
Ionosphere      & 143      & 2      & 351     & 316               & fico            & 10459   & 17       & 2  & 10459   & 3243                         \\
derm            & 131      & 6      & 358     & 358               & shuttleM        & 14500   & 198      & 2  & 14500   & 12978                        \\
dermatology     & 66       & 6      & 358     & 306               & eeg             & 14980   & 61       & 2  & 14980   & 5933                         \\
colic           & 415      & 2      & 368     & 357               & HTRU\_2         & 17898   & 53       & 2  & 17898   & 2234                         \\
balance         & 20       & 3      & 625     & 625               & magic04         & 19020   & 79       & 2  & 19020   & 10167                        \\
diabetes        & 112      & 2      & 768     & 768               & default\_credit & 30000   & 44       & 4  & 30000   & 18239                        \\
IndiansDiabetes & 11       & 2      & 768     & 175               & Adult           & 32561   & 81       & 2  & 32561   & 23874                        \\
anneal          & 84       & 2      & 812     & 489               & sepsis          & 110204  & 14       & 2  & 110204  & 140                          \\
vehicle         & 252      & 2      & 846     & 846               & skin            & 245057  & 80       & 2  & 245057  & 5466                        
\\ \bottomrule 
\end{tabular}
}
\end{threeparttable}
\label{uniquetsummary}
\end{table}

\section{Additional Metrics and Exact MIP Reformulations} 
\label{app:other-metrics} 
This appendix collects mixed-integer formulations for three additional performance metrics that are frequently used on imbalanced datasets: linear objectives  (balanced accuracy and class cost-sensitive loss)  and the nonlinear objectives (Geometric Mean,  Fowlkes--Mallows index, and the Jaccard/Intersection-over-Union (IoU)). 

\subsection{Linear Metrics} 
\label{subsec:linear-metrics} 
We collect four commonly used \emph{linear} objectives that are directly compatible with the formulation (\ref{eq:bendersOCT_master_2}). In all cases, one may optionally add a structural penalty (e.g., complexity) $-\lambda\sum_{n\in\mathcal B}\sum_{f\in\mathcal F} b_{nf}$ to the objective without affecting linearity. The structural/tree constraints (flow, split selection, leaf assignment, Benders cuts) are assumed given by Eqs.\ \eqref{eq:bendersOCT_master_2_benders_cut}--\eqref{eq:bendersOCT_master_2_g_upperbound}. \paragraph{Balanced Accuracy.} Balanced Accuracy (BA) averages per-class recalls: 

\begin{equation}
\mathrm{BA} = \frac{1}{2}\left(\frac{\mathrm{TP}}{n^+} + \frac{\mathrm{TN}}{n^-}\right) = \frac{1}{2}\left(\frac{\sum\limits_{i\in\mathcal{U}:~y_i=1} w_i\, g^i}{n^+} + \frac{\sum\limits_{i\in\mathcal{U}:~y_i=0} w_i\, g^i}{n^-}\right).  \end{equation} 

Because $n^+,n^-$ are data constants, maximizing BA is linear. Maximizing BA can be achieved by replacing the objective function of Problem (\ref{eq:bendersOCT_master_2}) with 
\begin{equation} \label{eq:BA} \max \quad \frac{1}{2}\left(\frac{\sum\limits_{i\in\mathcal{U}:~y_i=1} w_i\, g^i}{n^+} + \frac{\sum\limits_{i\in\mathcal{U}:~y_i=0} w_i\, g^i}{n^-}\right)- \lambda \displaystyle\sum_{n\in\mathcal B}\sum_{f\in\mathcal F} b_{nf}. \end{equation}

\paragraph{Class Cost-Sensitive Loss.} 
Given misclassification costs $c^+ \ge 0$ for false negatives and $c^- \ge 0$ for false positives, the total cost is 
\begin{equation} 
\mathrm{CS} = c^+\,\mathrm{FN}\;+\;c^-\,\mathrm{FP}= c^+\left({n^+} - {\sum\limits_{i\in\mathcal{U}:~y_i=1} w_i\, g^i}\right) + c^- (n^- -{\sum\limits_{i\in\mathcal{U}:~y_i=0} w_i\,g^i}). 
\end{equation} 
A linear minimization (or, equivalently, maximization of a linear reward) is \begin{equation} \label{eq:CS} \min\ \ \mathrm{CS}\;+\;\lambda\!\sum_{n,f} b_{nf} \quad\Longleftrightarrow\quad \max\ \ c^+\,\mathrm{TP} + c^-\,\mathrm{TN} \;-\; \lambda \displaystyle\sum_{n\in\mathcal B}\sum_{f\in\mathcal F} b_{nf}. \end{equation} 

\paragraph{Instance Cost-Sensitive Loss (minimize).} 
Let $\kappa_i\ge 0$ be an instance-specific penalty for misclassifying $i$. The instance-level cost is 
\[ \mathrm{ICS} \;=\; \sum_{i\in\mathcal{U}} \kappa_i\,w_i\,(1-g^i) \;=\; \Bigl(\sum_{i\in\mathcal{U}} \kappa_i w_i\Bigr) \;-\; \sum_{i\in\mathcal{U}} \kappa_i\,w_i\,g^i, \] so minimizing $\mathrm{ICS}$ is equivalent to maximizing the linear reward $\sum_i \kappa_i w_i g^i$. 
The corresponding linear program is 
\begin{equation} \label{eq:ICS} \min\ \ \sum_{i\in\mathcal{U}} \kappa_i\,w_i\,(1-g^i)\;+\;\lambda \displaystyle\sum_{n\in\mathcal B}\sum_{f\in\mathcal F} b_{nf} \quad\Longleftrightarrow\quad \max\ \ \sum_{i\in\mathcal{U}} \kappa_i\,w_i\,g^i\;-\lambda \displaystyle\sum_{n\in\mathcal B}\sum_{f\in\mathcal F} b_{nf}. \end{equation}

\medskip

\subsection{Squared Geometric Mean (G-Mean$^2$)} 
The G-Mean is defined as 
\[ \mathrm{G\text{-}Mean} \;=\; \sqrt{\frac{\mathrm{TP}}{\mathrm{TP}+\mathrm{FN}}\cdot \frac{\mathrm{TN}}{\mathrm{TN}+\mathrm{FP}}} \;=\; \sqrt{\frac{\mathrm{TP}\cdot \mathrm{TN}}{n^+ n^-}}\!, \] so it suffices to maximize $\mathrm{G2}=\mathrm{G\text{-}Mean}^2\in[0,1]$: \begin{equation} 
\label{app:eq:G2-MIQCP} 
\begin{array}{rlll} \max & \mathrm{G2} -\lambda \displaystyle\sum_{n\in\mathcal B}\sum_{f\in\mathcal F} b_{nf}\\[1pt] \text{s.t.} & \mathrm{G2}\cdot (n^+ n^-)\ \le\ \mathrm{TP}\cdot \mathrm{TN},\\ & \text{Constraints\ \eqref{eq:tp-fn}--\eqref{eq:tn-fp}, \eqref{eq:bendersOCT_master_2_benders_cut}--\eqref{eq:bendersOCT_master_2_g_upperbound}.} 
\end{array}
\end{equation} 

Let
\begin{align} \sum_{i\in\mathcal{U}:~y_i=1} w_i\, g^i &= \sum_{k=0}^{p} 2^{k}\,\gamma^+_k, & \sum_{i\in\mathcal{U}:~y_i=0} w_i\, g^i &= \sum_{l=0}^{q} 2^{l}\,\gamma^-_l. \label{app:eq:bit-consistency} \end{align} 

We use the following AND binaries throughout: 
\[ \eta_{kl} = \gamma^+_k \land \gamma^-_l \quad\Rightarrow\quad \begin{cases} \eta_{kl}\le \gamma^+_k,\ \eta_{kl}\le \gamma^-_l,\\ \eta_{kl}\ge \gamma^+_k+\gamma^-_l-1, \end{cases} \] \[ \xi^{++}_{rs} = \gamma^+_r \land \gamma^+_s \quad\Rightarrow\quad \begin{cases} \xi^{++}_{rs}\le \gamma^+_r,\ \xi^{++}_{rs}\le \gamma^+_s,\\ \xi^{++}_{rs}\ge \gamma^+_r+\gamma^+_s-1, \end{cases} \] so that \[ \mathrm{TP}\cdot\mathrm{TN}=\sum_{k=0}^p\sum_{l=0}^q 2^{k+l}\,\eta_{kl}, \qquad \mathrm{TP}^2=\sum_{r=0}^p\sum_{s=0}^p 2^{r+s}\,\xi^{++}_{rs}. \] 

Substitute the binary expansions of $\mathrm{TP}$ and $\mathrm{TN}$ and linearize $\gamma^+_k\gamma^-_l$ with $\eta_{kl}$. Model \eqref{app:eq:G2-MIQCP} is equivalent to the following MIP: 
\begin{equation} 
\label{app:eq:G2-MIP} 
\begin{array}{rlll} \max & \mathrm{G2}-\lambda \displaystyle\sum_{n\in\mathcal B}\sum_{f\in\mathcal F} b_{nf}\\[1pt] 
\text{s.t.} & \mathrm{G2}\cdot(n^+ n^-) \ \le\ \displaystyle\sum_{k=0}^p\sum_{l=0}^q 2^{k+l}\,\eta_{kl}\\[2pt] & \eta_{kl}\le \gamma^+_k,\ \eta_{kl}\le \gamma^-_l,\ \eta_{kl}\ge \gamma^+_k+\gamma^-_l-1 \quad \forall k,l,\\ & \text{\eqref{app:eq:bit-consistency} and \eqref{eq:bendersOCT_master_2_benders_cut}--\eqref{eq:bendersOCT_master_2_g_upperbound}.} \end{array} \end{equation} \subsection{Fowlkes--Mallows (squared)} The Fowlkes--Mallows index is \[ \mathrm{FM} \;=\; \sqrt{\frac{\mathrm{TP}}{\mathrm{TP}+\mathrm{FP}}\cdot \frac{\mathrm{TP}}{\mathrm{TP}+\mathrm{FN}}} \;=\; \sqrt{\frac{\mathrm{TP}^2}{n^+\,(n^- + \mathrm{TP}-\mathrm{TN})}}\,. \] Maximizing $\mathrm{FM}$ is equivalent to maximizing $\mathrm{FM2}:=\mathrm{FM}^2\in[0,1]$. \paragraph{MIQCP.} \begin{equation} \label{app:eq:FM2-MIQCP} \begin{array}{rlll} \max & \mathrm{FM2}\\[1pt] \text{s.t.} & \mathrm{FM2}\cdot\bigl[n^+\,(n^- + \mathrm{TP}-\mathrm{TN})\bigr] \ \le\ \mathrm{TP}^2,\\ & \text{Constraints\ \eqref{eq:tp-fn}--\eqref{eq:tn-fp}, \eqref{eq:bendersOCT_master_2_benders_cut}--\eqref{eq:bendersOCT_master_2_g_upperbound}.} \end{array} \end{equation} Write $n^- + \mathrm{TP}-\mathrm{TN} = n^- + \sum_k 2^k\gamma^+_k - \sum_l 2^l\gamma^-_l$ and linearize the products $\mathrm{FM2}\cdot \gamma^\pm$ via McCormick envelopes. Linearize $\mathrm{TP}^2$ with $\xi^{++}_{rs}$. Model \eqref{app:eq:FM2-MIQCP} is equivalent to: \begin{equation} \label{app:eq:FM2-MIP} \begin{array}{rlll} \max & \mathrm{FM2}\\[2pt] \text{s.t.} & \mathrm{FM2}\cdot (n^+ n^-) \;+\; n^+\!\left(\displaystyle\sum_{k=0}^{p} 2^k\,\delta^{+}_k \;-\; \sum_{l=0}^{q} 2^l\,\delta^{-}_l\right) \ \le\ \displaystyle\sum_{r=0}^{p}\sum_{s=0}^{p} 2^{r+s}\,\xi^{++}_{rs} \\[6pt] & \delta^{+}_k = \mathrm{FM2}\cdot \gamma^{+}_k \ \text{via}\ \begin{cases} \delta^{+}_k \le \gamma^{+}_k,\\ \delta^{+}_k \le \mathrm{FM2},\\ \delta^{+}_k \ge \mathrm{FM2} - (1-\gamma^{+}_k), \end{cases} \quad \forall k,\\[8pt] & \delta^{-}_l = \mathrm{FM2}\cdot \gamma^{-}_l \ \text{via}\ \begin{cases} \delta^{-}_l \le \gamma^{-}_l,\\ \delta^{-}_l \le \mathrm{FM2},\\ \delta^{-}_l \ge \mathrm{FM2} - (1-\gamma^{-}_l), \end{cases} \quad \forall l,\\[10pt] & \xi^{++}_{rs} = \gamma^{+}_r \land \gamma^{+}_s \ \text{via}\ \begin{cases} \xi^{++}_{rs}\le \gamma^{+}_r,\\ \xi^{++}_{rs}\le \gamma^{+}_s,\\ \xi^{++}_{rs}\ge \gamma^{+}_r+\gamma^{+}_s-1, \end{cases} \quad \forall r,s,\\[10pt] & \text{\eqref{app:eq:bit-consistency} and \eqref{eq:bendersOCT_master_2_benders_cut}--\eqref{eq:bendersOCT_master_2_g_upperbound}.} \end{array} \end{equation} 

\subsection{Jaccard / Intersection-over-Union (IoU)} 
The IoU (also called Jaccard index) is 
\[ \mathrm{IoU} \;=\; \frac{\mathrm{TP}}{\mathrm{TP}+\mathrm{FP}+\mathrm{FN}} \;=\; \frac{\mathrm{TP}}{|\mathcal I|-\mathrm{TN}}\,, \qquad \mathrm{IoU}\in[0,1]. \] \paragraph{MIQCP.} \begin{equation} \label{app:eq:IoU-MIQCP} \begin{array}{rlll} \max & \mathrm{IoU}\\[1pt] \text{s.t.} & \mathrm{IoU}\cdot (|\mathcal I|-\mathrm{TN}) \ \le\ \mathrm{TP},\\ & \text{Constraints\ \eqref{eq:tp-fn}--\eqref{eq:tn-fp}, \eqref{eq:bendersOCT_master_2_benders_cut}--\eqref{eq:bendersOCT_master_2_g_upperbound}.} 
\end{array} 
\end{equation} 

Expand $|\mathcal I|-\mathrm{TN}=|\mathcal I|- \sum_l 2^l\gamma^-_l$ and linearize $\mathrm{IoU}\cdot \gamma^-_l$ with McCormick envelopes. Model \eqref{app:eq:IoU-MIQCP} is equivalent to: 
\begin{equation} 
\label{app:eq:IoU-MIP} 
\begin{array}{rlll} 
\max & \mathrm{IoU} - \lambda \displaystyle\sum_{n\in\mathcal B}\sum_{f\in\mathcal F} b_{nf}\\[2pt] \text{s.t.} & \mathrm{IoU}\cdot |\mathcal I| \;-\; \displaystyle\sum_{l=0}^{q} 2^l \,\delta^{-}_l \ \le\ \displaystyle\sum_{i\in\mathcal U: y_i=1} w_i g^i\\[6pt] & \delta^{-}_l = \mathrm{IoU}\cdot \gamma^{-}_l \ \text{via}\ \begin{cases} \delta^{-}_l \le \gamma^{-}_l,\\ \delta^{-}_l \le \mathrm{IoU},\\ \delta^{-}_l \ge \mathrm{IoU} - (1-\gamma^{-}_l), \end{cases} \quad \forall l,\\[10pt] & \text{\eqref{app:eq:bit-consistency} and \eqref{eq:bendersOCT_master_2_benders_cut}--\eqref{eq:bendersOCT_master_2_g_upperbound}.} 
\end{array} 
\end{equation} 

\subsection{Diagnostic Odds Ratio (DOR)} 
The diagnostic odds ratio is 
\[ \mathrm{DOR} \;=\; \frac{\mathrm{TP}\cdot \mathrm{TN}}{\mathrm{FP}\cdot \mathrm{FN}} \;=\; \frac{\mathrm{TP}\cdot \mathrm{TN}}{n^+ n^- - n^+\,\mathrm{TN} - n^-\,\mathrm{TP} + \mathrm{TP}\cdot \mathrm{TN}}\,. \] 

Because $\mathrm{DOR}$ can be unbounded when $\mathrm{FP}$ or $\mathrm{FN}$ equals $0$, an exact MIP needs either (i) a modeling upper bound $\overline D$ on $\mathrm{DOR}$, or (ii) a small denominator smoothing $\varepsilon>0$. We provide both variants. \paragraph{Variant A (bounded DOR).} 
Assume a user-specified bound $\overline D>0$ (e.g., from validation or domain knowledge) and constrain $0\le \mathrm{DOR}\le \overline D$. 

\begin{equation} 
\label{app:eq:DOR-MIQCP} 
\begin{array}{rlll} \max & \mathrm{DOR}-\lambda \displaystyle\sum_{n\in\mathcal B}\sum_{f\in\mathcal F} b_{nf}\\[2pt] \text{s.t.} & \mathrm{DOR}\,\Bigl[n^+ n^- - n^+\,\mathrm{TN} - n^-\,\mathrm{TP} + \mathrm{TP}\cdot \mathrm{TN}\Bigr] \ \le\ \mathrm{TP}\cdot \mathrm{TN},\\ & 0\le \mathrm{DOR}\le \overline D,\\ & \text{Constraints\ \eqref{eq:tp-fn}--\eqref{eq:tn-fp}, \eqref{eq:bendersOCT_master_2_benders_cut}--\eqref{eq:bendersOCT_master_2_g_upperbound}.} 
\end{array} 
\end{equation} 

Expand the bracketed term linearly in $\gamma^\pm$ and $\eta_{kl}$; linearize the products $\mathrm{DOR}\cdot \gamma^\pm$ and $\mathrm{DOR}\cdot \eta_{kl}$ via McCormick with upper bound $\overline D$. Under $0\le \mathrm{DOR}\le \overline D$, \eqref{app:eq:DOR-MIQCP} is equivalent to: 
\begin{equation} 
\label{app:eq:DOR-MIP} 
\begin{array}{rlll} 
\max & \mathrm{DOR}-\lambda \displaystyle\sum_{n\in\mathcal B}\sum_{f\in\mathcal F} b_{nf}\\[2pt] 
\text{s.t.} & \mathrm{DOR}\cdot n^+ n^- \;-\; n^+\sum\limits_{l=0}^{q} 2^l\,\delta^{-}_l \;-\; n^-\sum\limits_{k=0}^{p} 2^k\,\delta^{+}_k \;+\; \sum\limits_{k=0}^{p}\sum\limits_{l=0}^{q} 2^{k+l}\,\zeta_{kl} \le\ \sum\limits_{k=0}^{p}\sum\limits_{l=0}^{q} 2^{k+l}\,\eta_{kl}\\[6pt] 
& \delta^{+}_k = \mathrm{DOR}\cdot \gamma^{+}_k \ \text{via}\ \begin{cases} \delta^{+}_k \le \overline D\,\gamma^{+}_k,\\ \delta^{+}_k \le \mathrm{DOR},\\ \delta^{+}_k \ge \mathrm{DOR} - \overline D(1-\gamma^{+}_k), \end{cases} \quad \forall k,\\[10pt] & \delta^{-}_l = \mathrm{DOR}\cdot \gamma^{-}_l \ \text{via}\ \begin{cases} \delta^{-}_l \le \overline D\,\gamma^{-}_l,\\ \delta^{-}_l \le \mathrm{DOR},\\ \delta^{-}_l \ge \mathrm{DOR} - \overline D(1-\gamma^{-}_l), \end{cases} \quad \forall l,\\[10pt] & \zeta_{kl} = \mathrm{DOR}\cdot \eta_{kl} \ \text{via}\ \begin{cases} \zeta_{kl} \le \overline D\,\eta_{kl},\\ \zeta_{kl} \le \mathrm{DOR},\\ \zeta_{kl} \ge \mathrm{DOR} - \overline D(1-\eta_{kl}), \end{cases} \quad \forall k,l,\\[10pt] & \eta_{kl} = \gamma^+_k \land \gamma^-_l \ \text{as above},\\ & \text{\eqref{app:eq:bit-consistency} and \eqref{eq:bendersOCT_master_2_benders_cut}--\eqref{eq:bendersOCT_master_2_g_upperbound}.} \end{array} 
\end{equation}

\end{document}